\pdfoutput=1

\documentclass[11pt]{article}

\usepackage{acl}

\usepackage{times}
\usepackage{latexsym}

\usepackage[T1]{fontenc}

\usepackage[utf8]{inputenc}

\usepackage{microtype}

\usepackage{inconsolata}

\usepackage{graphicx}
\usepackage{amssymb}
\usepackage{enumitem}
\usepackage{booktabs}
\usepackage{pifont}
\usepackage{fontawesome}
\usepackage{multirow}
\usepackage{amsmath}
\usepackage{soul}
\usepackage{subcaption}

\usepackage{booktabs, adjustbox, array}

\title{Time-MQA: Time Series Multi-Task Question Answering with \\ Context Enhancement}

\author{
 \textbf{Yaxuan Kong\textsuperscript{1}\footnotemark[1]},
 \textbf{Yiyuan Yang\textsuperscript{1,2}\footnotemark[1]},
 \textbf{Yoontae Hwang\textsuperscript{1}},
 \textbf{Wenjie Du\textsuperscript{2}},
 \textbf{Stefan Zohren\textsuperscript{1}},
\\
 \textbf{Zhangyang Wang\textsuperscript{3}},
 \textbf{Ming Jin\textsuperscript{4}\footnotemark[2]},
 \textbf{Qingsong Wen\textsuperscript{1,5}\footnotemark[2]},
\\
 \textsuperscript{1}University of Oxford,
 \textsuperscript{2}PyPOTS Research,
 \textsuperscript{3}University of Texas at Austin,
\\
 \textsuperscript{4}Griffith University,
 \textsuperscript{5}Squirrel Ai Learning
\\
 \small{
   \href{mailto:email@domain}{yaxuan.kong@eng.ox.ac.uk},
   \href{mailto:email@domain}{yiyuan.yang@cs.ox.ac.uk}
 }
}

\begin{document}
\maketitle
\renewcommand{\thefootnote}{\fnsymbol{footnote}}
\footnotetext[1]{Equal Contribution.}
\footnotetext[2]{Corresponding Authors.}
\renewcommand{\thefootnote}{\arabic{footnote}}
\begin{abstract}
Time series data are foundational in finance, healthcare, and energy domains. However, most existing methods and datasets remain focused on a narrow spectrum of tasks, such as forecasting or anomaly detection. To bridge this gap, we introduce Time Series Multi-Task Question Answering (Time-MQA), a unified framework that enables natural language queries across multiple time series tasks - numerical analytical tasks and open-ended question answering with reasoning. Central to Time-MQA is the TSQA dataset, a large-scale dataset containing $\sim$200k question-answer pairs derived from diverse time series spanning environment, traffic, etc. This comprehensive resource covers various time series lengths and promotes robust model development. We further demonstrate how continually pre-training large language models (Mistral 7B, Llama-3 8B, and Qwen-2.5 7B) on the TSQA dataset enhanced time series reasoning capabilities, moving beyond mere numeric tasks and enabling more advanced and intuitive interactions with temporal data. The complete TSQA dataset, models, user study questionnaires for evaluation, and other related materials have been open-sourced here\footnote{https://huggingface.co/Time-MQA}.
\end{abstract}

\section{Introduction}
\begin{figure}[!t]
\begin{center}
\includegraphics[width = 1\linewidth]{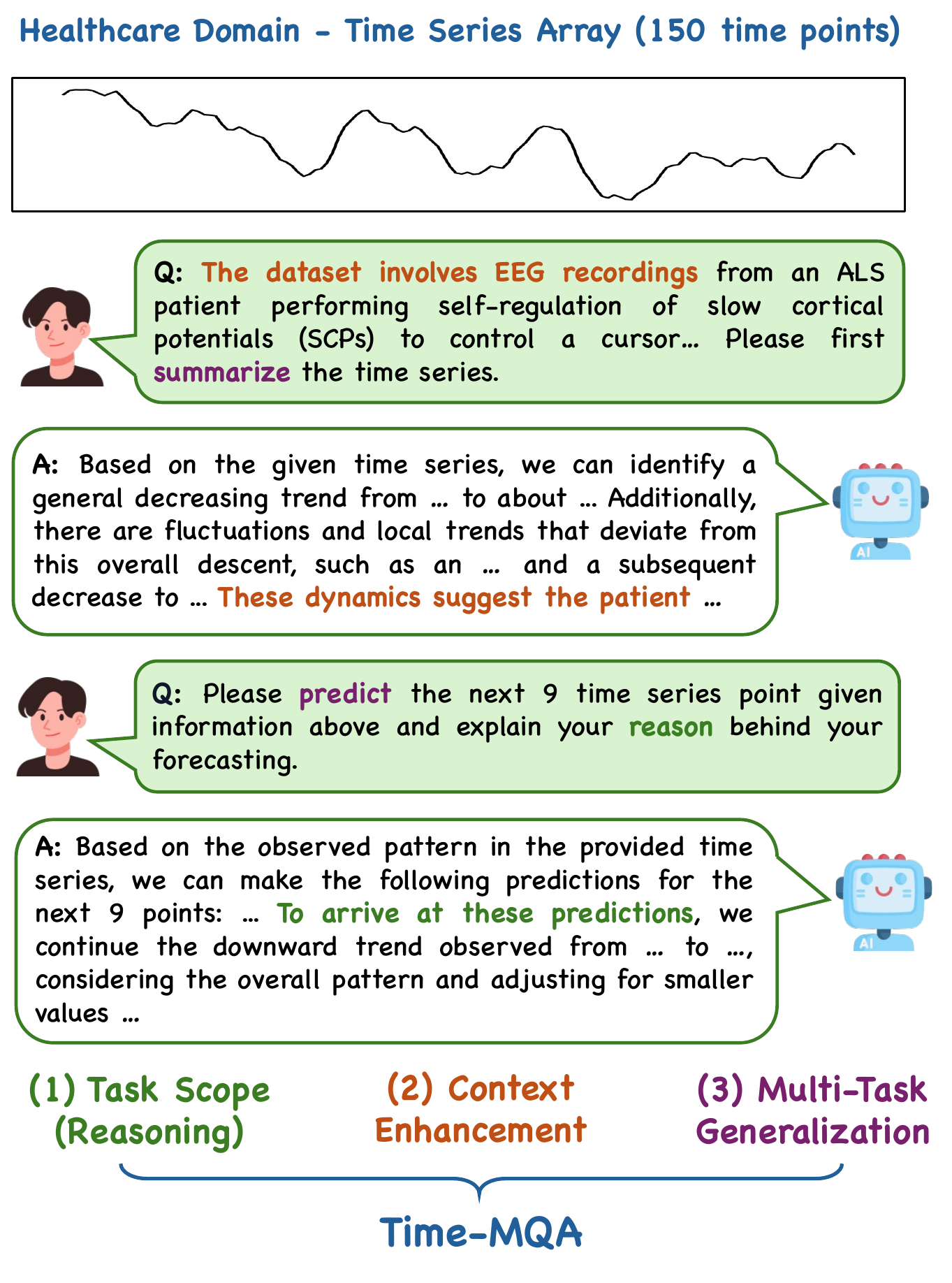}
\end{center}
\caption{Example of Time-MQA with context enhancement. The LLM answers the user’s questions based on the provided context and the input time series. It can conduct reasoning and handle multiple time series tasks.} 
\vspace{-10pt}
\label{Figure_1}
\end{figure}

Time series analysis has long been fundamental to real-world applications in finance, healthcare, energy, and other domains~\cite{nie2024survey,xu2023density,yang2021long,chen2024deep}. Before the emergence of large language models (LLMs), most research in this area focused on numerical analytical tasks such as forecasting and anomaly detection, with methods typically constrained to a single objective. Recently, the community has begun to explore ways to enhance these conventional time series approaches by incorporating LLMs~\cite{fons2024evaluating,zhang2024dualtime}; however, while some efforts do leverage contextual information, they typically focus on a single task, such as forecasting, leaving a gap in broader, multi-task reasoning and inference capabilities~\cite{merrill2024language,ansari2024chronos,frisoni2024generate,jin2024mmtom}. To bridge this gap, we propose a unified \emph{Time Series Multi-task Question Answering (Time-MQA)} framework that integrates diverse tasks with natural language queries (Figure \ref{Figure_1}).

\textit{Time-MQA} is crucial for advanced reasoning and inference, as it enables models to interpret temporal data through natural language queries and uncover deeper insights beyond mere classical numeric tasks~\cite{chow2024towards,xu2024beyond}. However, there is a notable lack of relevant datasets — specifically, paired language with time series — which severely limits the development of models capable of dialogue and reasoning in this domain~\cite{jin2024position,kong2025position}. This significant gap underscores the urgent need for comprehensive resources that integrate multiple time series tasks under a question-answering framework, allowing users to query these tasks in natural language.

In this paper, we introduce \textit{Time-MQA}, a new framework for multi-task time series question answering with context enhancement. As illustrated in Figure \ref{Figure_1}, \textit{Time-MQA} expands beyond numerical analytical tasks by consolidating open-ended questions and classical time series tasks, such as forecasting, into a single framework. Users can pose queries in natural language, enabling a more intuitive and flexible interface for a wide range of time series analyses.

To support \textit{Time-MQA}, we proposed \textit{TSQA}, a large-scale dataset with approximately 200k question-answer pairs spanning multiple domains, including healthcare, environment, AIOps, machine sensors, finance, energy, traffic, IoT, nature, transport, human activities, and the web. This dataset covers various time series lengths and tasks, ensuring broad coverage and robustness. Notably, \textit{TSQA} features open-ended reasoning questions with more elaborate text-based explanations. To the best of our knowledge, this is the first large-scale QA dataset in the time series domain that spans multiple domains and tasks, effectively bridging the gap between classical time series analysis and modern LLM-driven approaches.

We further demonstrate the utility of the \textit{TSQA} dataset by employing continual pre-training techniques on representative LLMs such as Mistral 7B \cite{jiang2023mistral}, Llama-3 8B \cite{dubey2024llama}, and Qwen-2.5 7B \cite{yang2024qwen2}. Our experiments reveal that these models trained on the \textit{TSQA} dataset can effectively acquire time series knowledge and reasoning abilities, enabling more advanced capabilities beyond basic numeric handling of time series data.

In summary, our contributions include: 
\begin{itemize} [leftmargin=1em, itemsep=0.2em, parsep=0em, topsep=0.3em]
    \item We propose \textit{Time-MQA}, a multi-task time series question answering framework that leverages contextual enhancement to extend beyond traditional numerical analysis tasks.
    \item We construct \textit{TSQA}, a dataset comprising $\sim$200k question-answer pairs across over twelve domains (e.g., healthcare, finance, and energy) and five tasks (e.g., forecasting, anomaly detection, and open-ended reasoning question answering).
    \item We demonstrate that fine-tuning LLMs on the \textit{TSQA} dataset equips them with time series–specific knowledge, enabling natural language queries for comprehensive time series analysis.
\end{itemize}

\begin{figure*}[!t]
\begin{center}
\includegraphics[width = 1\linewidth]{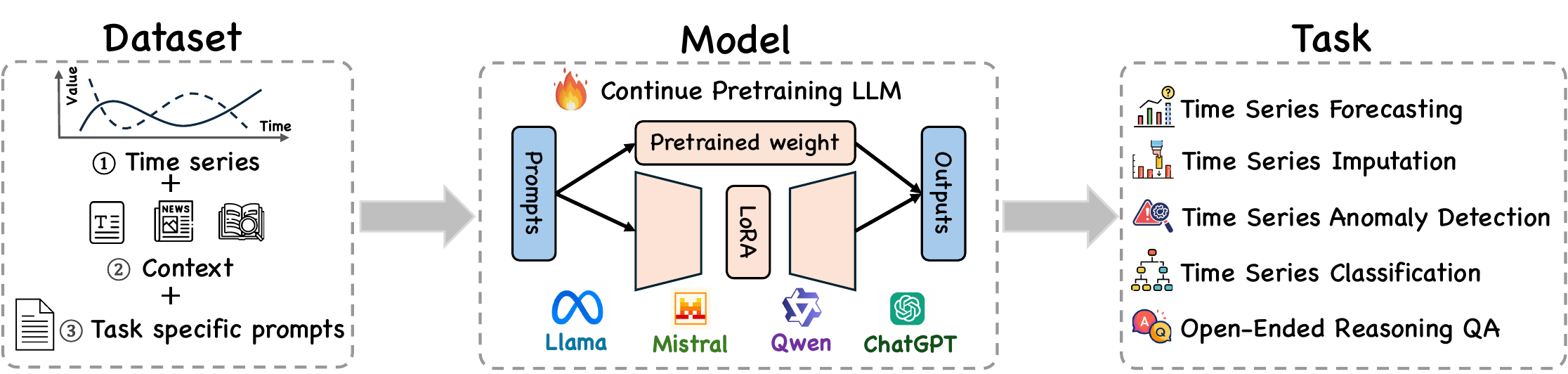}
\end{center}
\caption{The overview of the proposed Time-MQA framework.}
\vspace{-10pt}
\label{Figure_framework}
\end{figure*}

\section{Related Works}
\subsection{Classical Time Series Tasks} 
Time series analysis has been extensively studied in various real-world applications, such as finance, healthcare, climate, electricity, AIOps, and industrial system maintenance~\cite{nie2024survey,guo2024maximizing,ma2024fusionsf,yang2023sgdp,yang2021early}. Classical time series tasks focus on extracting insights from these time series and addressing challenges associated with temporal patterns~\cite{fuller2009introduction,hamilton2020time}. 

\textbf{Forecasting} is one of the most fundamental tasks. It predicts future time points based on historical values and features~\cite{lim2021time}. Depending on the temporal horizon, forecasting can be categorized into short-term forecasting, which captures immediate fluctuations, and long-term forecasting, which models more complex temporal dependencies~\cite{wang2024timexer}. Common methods include statistical methods (e.g., ARIMA and exponential smoothing) and deep learning-based methods (e.g., RNN, LSTM, Transformer-based architectures)~\cite{wen2022transformers,miller2024survey}. 

\textbf{Anomaly detection} seeks to identify abnormal patterns or deviations from expected behavior in time-series data~\cite{zamanzadeh2024deep}. Classical approaches rely on statistical models like z-score analysis and dynamic thresholding, while contemporary methods incorporate deep learning-based frameworks, such as autoencoders, transformers, and graph neural networks, to capture complex dependencies and temporal correlations~\cite{han2022adbench,yang2023dcdetector}. 

\textbf{Imputation} addresses the issue of missing or corrupted data in sequence. It is essential for ensuring data integrity in downstream analysis~\cite{du2024tsi}. Traditional imputation techniques, such as interpolation, have been widely employed, whereas deep learning-based methods, such as variational autoencoders (VAE), generative adversarial networks (GANs), and diffusion models, have recently demonstrated promising results in learning complex missing patterns and improving imputation performance~\cite{wang2025deeplearningmultivariatetime}. 

Beyond these tasks, time series classification/regression, generation, augmentation, and decomposition are also frequently used in real-world scenarios~\cite{mohammadi2024deep,wen2020time,zhang2024self}.

\subsection{Text-Enhanced Time Series Tasks} 
Recent advancements in time series analysis have demonstrated the potential of incorporating textual information based on LLMs to enhance time series tasks~\cite{jin2024position,wang2024chattimeunifiedmultimodaltime}. Unlike classical approaches that rely solely on numerical data, text-enhanced time series analysis leverages domain-specific textual descriptions, contextual metadata, or associated reports to improve the cognitive understanding and modeling of time-dependent patterns~\cite{liu-etal-2024-lstprompt}. This hybrid approach mitigates the limitations of unimodal time series models by integrating additional semantic and contextual cues that are useful for decision-making~\cite{kong2025position,singh2024finqapt}. 

In detail, text-enhanced time series forecasting and anomaly detection tasks benefit from textual information by incorporating expert reports to refine predictions and provide anomaly causal explanations~\cite{hollmann2025accurate,chen2023tele}. Similarly, classification, imputation, and generation tasks can be enhanced by leveraging textual descriptions as auxiliary supervision or describing missing values~\cite{bernardini2023novel,moor2023foundation}. It will help models distinguish subtle variations across different categories and generate more informed reconstructions.

Recent research has explored various techniques for integrating textual and time-series data, including LLM-based alignment, cross-modal attention mechanisms, and contrastive learning strategies that jointly encode text and time-series representations~\cite{jin2023large,zhang2023insight,liu2024picture}. Some approaches, such as Time-LLM~\cite{jin2023time}, directly adapt LLMs to process text and time-series data, whereas others, like Time-MMD~\cite{liutime}, employ weighted fusion methods to combine textual embeddings with deep time-series backbones. By enhancing classical time series tasks with textual information, text-enhanced time series models offer greater robustness and richer interpretability across diverse applications~\cite{jin2024position,kong2025position}. 

\subsection{Language Question Answering}
Question Answering (QA) in Natural Language Processing (NLP) involves systems that interpret human language queries to retrieve or generate accurate answers~\cite{biancofiore2024interactive, chen2024spiral}. It evolves from rule-based systems to neural architectures driven by LLMs like GPT-4~\cite{achiam2023gpt} and Llama~\cite{touvron2023llama}. These models leverage massive text corpora and large-scale datasets for end-to-end pre-training, fine-tuned via supervised learning or reinforcement learning with human feedback (RLHF) to align responses with factual and contextual relevance~\cite{liu2023summary}. Innovations such as retrieval-augmented generation (RAG) combine parametric knowledge with external data sources, while benchmarks like SQuAD, HotpotQA, MuSiQue, FinTextQA, SyllabusQA, and ToolQA drive progress~\cite{trivedi2022musique, ho2020constructing,yang2018hotpotqa,zhuang2024toolqa,chen2024fintextqa,fernandez2024syllabusqa}. However, challenges remain, such as handling ambiguous queries, ensuring the accuracy of generated answers, and maintaining efficiency in processing large volumes of data. Ongoing research focuses on enhancing the reasoning capabilities of QA systems, improving their ability to handle complex and nuanced questions, and expanding their applicability across diverse domains~\cite{singh2025agentic}.

\begin{figure*}[!t]
\begin{center}
\includegraphics[width = 1\linewidth]{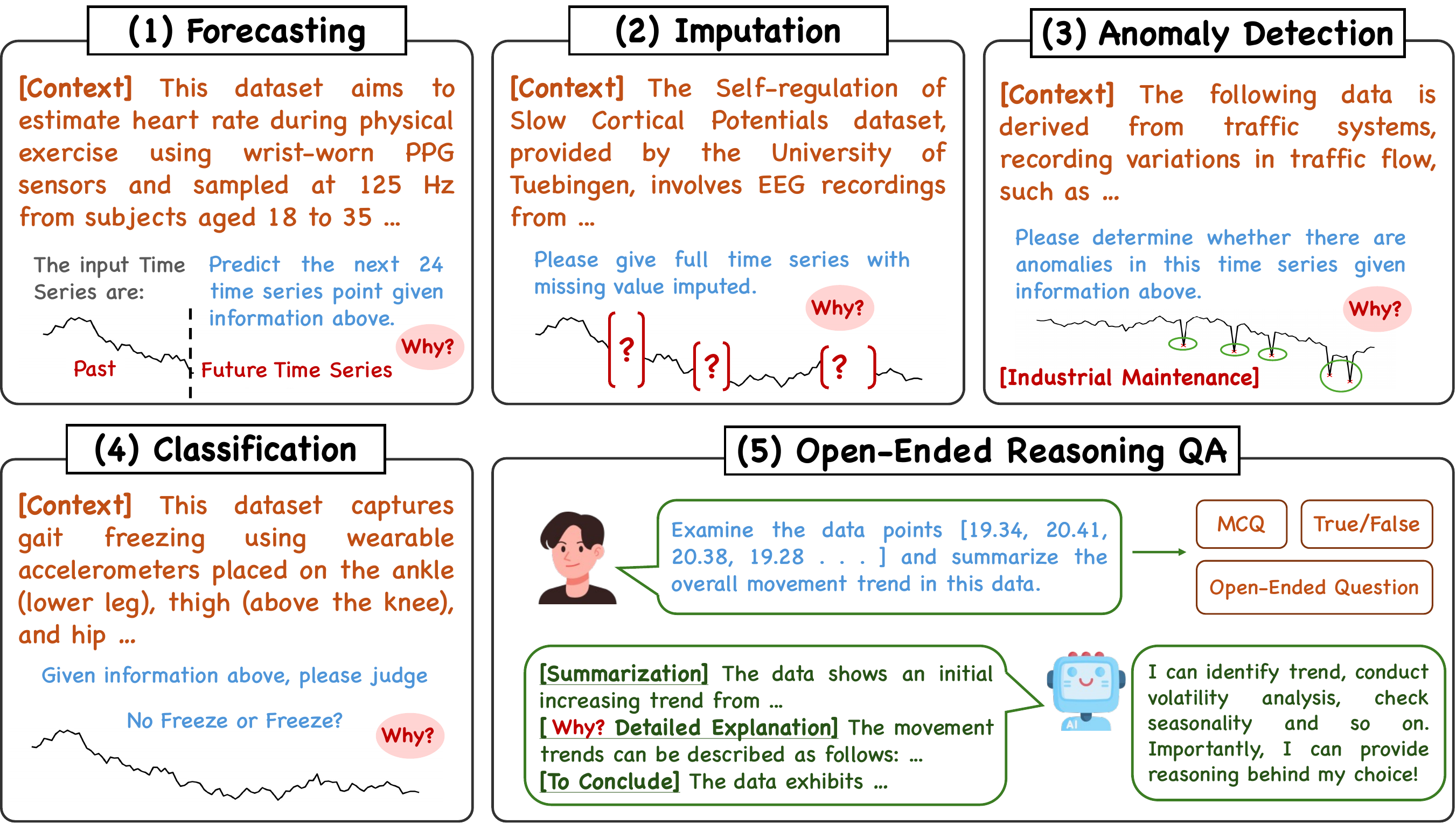}
\end{center}
\caption{The demonstration of the Time-MQA with context enhancement.}
\label{Figure_3}
\end{figure*}

\section{Methodology}
\subsection{The Time-MQA Framework}
The Time-MQA framework broadens traditional analysis by integrating diverse objectives - such as forecasting, imputation, classification, anomaly detection, and notably, open-ended queries - into a unified question-answering paradigm (as shown in Figure \ref{Figure_framework}). Formally, let $\mathbf{X} = \{x_1, x_2, ..., x_T\}$ denote a time series input, where $x_t \in \mathbb{R}^d$ represents a $d$-dimensional observation at timestep $t$. Let $\mathbf{C}$ represent additional contextual information (e.g., textual metadata, domain-specific knowledge, or other modalities). For a question $Q$ expressed in natural language, the goal is to generate an answer $A$ conditioned on both $X$ and contextual information $C$. Time-MQA aims to learn a function $f: (\mathbf{X}, \mathbf{C}, Q) \rightarrow A,$ where $A$ is the correct answer to the query $Q$. Depending on the nature of $Q$, $A$ can take diverse forms, such as predicted values, classification labels, a set of anomalous timestamps, and textual explanations. 
Specifically, the model component of Time-MQA is based on continued pre-trained LLMs (i.e., Mistral 7B \cite{jiang2023mistral}, Llama-3 8B \cite{dubey2024llama}, and Qwen-2.5 7B \cite{yang2024qwen2}), using the prepared TSQA dataset. To optimize parameter usage, Time-MQA employs Parameter-Efficient Fine-Tuning (PEFT) with a LoRA adapter. 
The supported various tasks of the Time-MQA with context enhancement are demonstrated with details in Figure ~\ref{Figure_3}.

\paragraph{Key Distinctions.}
There are three main differences between Time-MQA and traditional time series analysis tasks: 
\begin{enumerate}[label=(\textbf{\arabic*)}, itemsep=0.3em, parsep=0em, topsep=0.3em]
    \item \textbf{Task Scope:} Traditional tasks focus on singular objectives (e.g., forecasting future values or classifying the time series). In contrast, Time-MQA unifies these under a question-driven paradigm, enabling both conventional tasks (e.g., “Forecast the next 5 values”) and complex queries (e.g., “Why did the temperature drop abruptly at hour 10?”) that require joint reasoning across detection, explanation, and contextual knowledge.
    \item \textbf{Context Enhancement:} Traditional methods rely solely on the time series $\mathbf{X}$. Time-MQA integrates auxiliary context $\mathbf{C}$ to resolve ambiguities and improve robustness. For instance, even when analyzing the same time series, differences in a dataset’s background information can yield contrasting predictions \cite{williams2024context}.
    \item \textbf{Multi-Task Generalization:} Unlike single-purpose models, Time-MQA dynamically adapts to diverse question types through a unified architecture, eliminating the need for task-specific pipelines. This flexibility allows it to handle various time series tasks within a single framework, promoting knowledge sharing via shared representations that enhance both performance and interpretability.
\end{enumerate}

\begin{figure}[!t]
\begin{center}
\includegraphics[width = 0.9\linewidth]{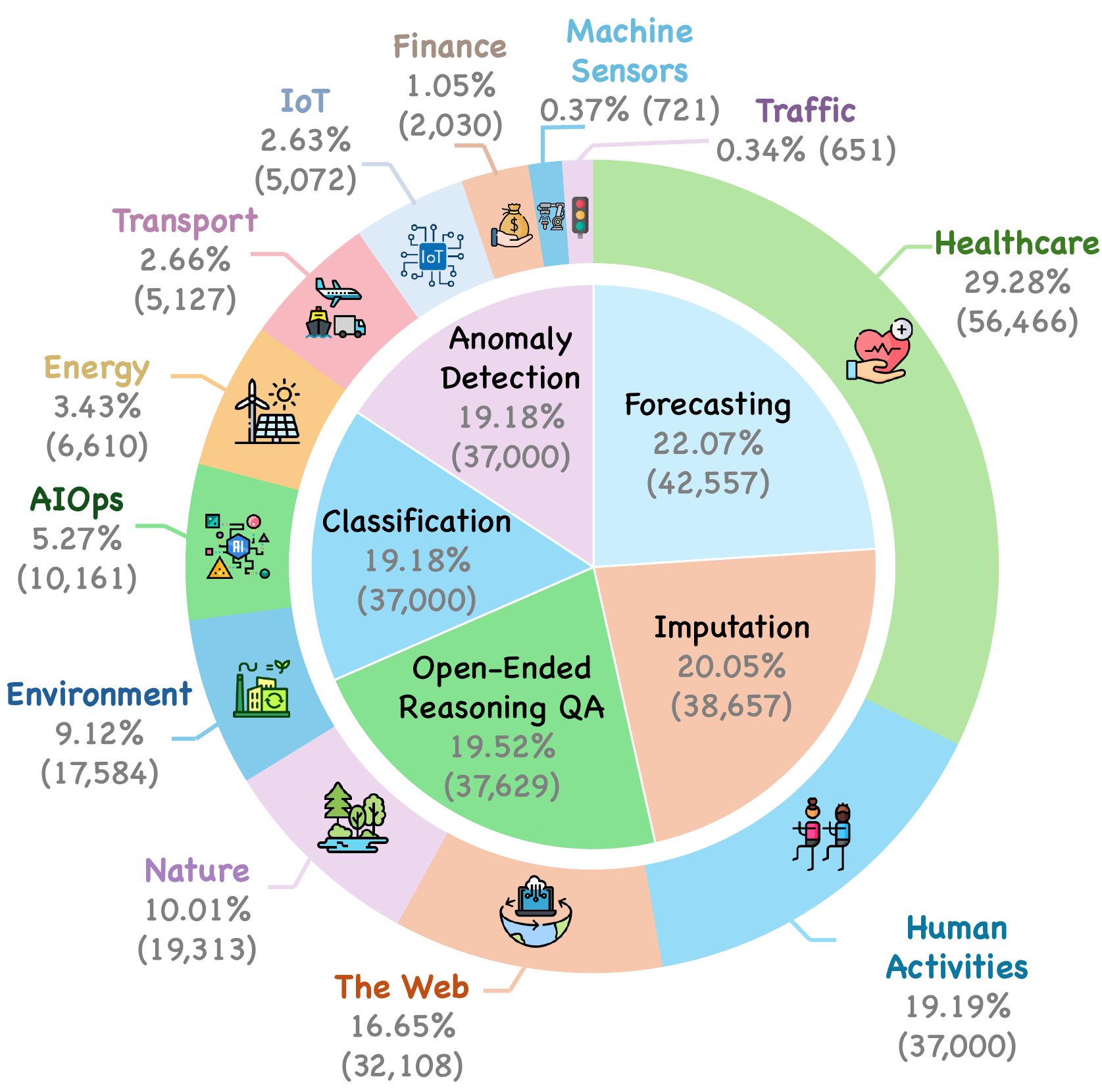}
\end{center}
\caption{The distribution of data statistics in the TSQA dataset. The inner ring shows task types, and the outer ring shows domains. The rings are presented together for comparison, with no direct one-to-one correspondence between their segments.}
\vspace{-5mm}
\label{Figure_4}
\end{figure}

\subsection{The TSQA Dataset}
This section will introduce our self-constructed 200k-level text-enhanced time series dataset, i.e., TSQA. From perspectives such as the data collection process and types across multiple domains and tasks. We compare it with other datasets and demonstrate the advantages of TSQA.

\subsubsection{Dataset Composition and Categorization} 
The raw data we use comes from various classic publicly available datasets in the time series domain, covering multiple tasks and application areas. Innovatively, we have incorporated textual descriptions, including background information, feature descriptions, etc. From a task perspective, our dataset can be categorized into the following types. Some examples of the proposed TSQA dataset can be found in the Appendix.

\paragraph{Forecasting.} In the forecasting task, we utilize UTSD datasets~\cite{liu2024timer}, publicly available time series forecasting datasets, such as ETTh1, ETTh2, ETTm1, ECL, Weather, etc.~\cite{zhou2021informer}, and the latest text-enhanced time series datasets as the raw time-series data \cite{liutime}. Additionally, for the first time, we incorporate financial datasets with earnings call transcripts~\cite{FoolEarnings2025}. To ensure the generalization ability of algorithms and models, we set the input length and prediction length to random values between 64–256 and 8–32, respectively. Furthermore, we enhance all data with background information, feature descriptions, and task descriptions as textual information based on the original data sources. In summary, the forecasting dataset includes 42,557 data instances spanning application domains such as energy, environment, health, IoT, nature, transport, Web, AIOps, etc.

\paragraph{Imputation.} The imputation task shares the same original time series data sources as the forecasting task. Additionally, we randomly set the input length between 64 and 256, randomly removed 4 to 12 values, and replaced them with "X". The imputation dataset consists of a total of 38,657 instances from multiple application domains.

\paragraph{Anomaly Detection.} For the anomaly detection task, we utilize commonly used public datasets in the field, such as UCR, ECG, KPI, MGAB, NAB, SensorScope, and Yahoo, among others~\cite{zamanzadeh2024deep}. To enhance the generalization capability of algorithms and models, we randomly set the input length between 8 and 256 while ensuring a balanced distribution between anomalous and normal data. The anomaly detection dataset comprises 37,000 instances from various application domains, including AIOps, Health, Finance, Machinery, Industrial Sensors, Environment, and Traffic.

\paragraph{Classification.} For the classification task, the data sources are relatively straightforward, primarily derived from the human activity recognition application domain~\cite{kwapisz2011activity,bachlin2009potentials}. The dataset includes both binary and multi-class classification tasks. To ensure class balance, we collected a total of 37,000 data instances. Additionally, the input sequence length is randomly set between 8 and 32 to maintain diversity in data representation.

\paragraph{Open-Ended Reasoning QA.}
We used different parts of the UTSD dataset~\cite{liu2024timer} - distinct from those employed in our forecasting tasks - to avoid data leakage. To generate open-ended reasoning QA, we utilized GPT-4o, instructing it to create questions covering various topics such as trends, seasonality, cyclical patterns, summarization, volatility, anomalies, structural breaks, and other statistical properties (without limiting it strictly to these areas). We also incorporated multiple question types, including multiple-choice, true/false, and open-ended formats. The prompts used to generate these QA pairs are provided in the Appendix. In total, we generated 37,629 data instances, from which we manually reviewed and selected 1,400 QA pairs for use in the continual pretraining step.

\subsubsection{Data Statistics}
The TSQA dataset comprises 192,843 ($\sim$200k) question-answer pairs spanning twelve domains - healthcare, finance, energy, traffic, environment, IoT, nature, transport, human activities, machine sensors, AIOps, and the web - and five task types: forecasting, imputation, anomaly detection, classification, and open-ended reasoning (see Figure \ref{Figure_4} for distributions). Within the open-ended reasoning QA, the dataset includes 6,919 true/false questions, 11,281 multiple-choice questions, and 12,510 open-ended questions, offering a broad and diverse range of question formats.

\begin{table}[!t]
\caption{Comparison of text-enhanced time series datasets. \faCheckCircleO ~and \ding{52} indicate having only real data and having both synthetic and real data.}
\resizebox{1\linewidth}{!}{
\fontsize{11}{14}\selectfont
\renewcommand{\arraystretch}{1}{
\setlength{\heavyrulewidth}{1.25pt} 
\begin{tabular}{lccccc}
\toprule   
\multicolumn{1}{l}{\textbf{Dataset}} & \multicolumn{1}{c}{\textbf{Numerical}} & \multicolumn{1}{c}{\textbf{Cognitive}} & \multicolumn{1}{c}{\textbf{Source}} & \multicolumn{1}{c}{\textbf{Domain}} & \multicolumn{1}{c}{\textbf{Size}} \\ \midrule
TS-Insights & \ding{52} & \ding{56} & \faCheckCircleO & 4 & \textasciitilde10k \\
ChatTS & \ding{56} & \ding{52} & \ding{52} & 4 & \textasciitilde2.2k \\
CiK & \ding{52} & \ding{56} & \faCheckCircleO & 7 & \textasciitilde2.9k\\
TimeMMD & \ding{52} & \ding{56} & \faCheckCircleO & 9 & \textasciitilde16k\\
\textbf{TSQA} & \ding{52} & \ding{52} & \ding{52} & \textbf{12} & \textasciitilde\textbf{200k}\\ \bottomrule
\end{tabular}}}
\label{tab:dataset}
\end{table}

\subsubsection{Comparison with Existing Datasets}

We summarize several existing datasets compared to our proposed TSQA dataset in Table~\ref{tab:dataset}. Unlike prior datasets (TS-Insights~\cite{zhang2023insight}, ChatTS~\cite{xie2024chatts}, CiK~\cite{williams2024context}, and TimeMMD~\cite{liutime}), which focus on either classical numerical analytical time series tasks (e.g., forecasting, anomaly detection, imputation, classification) or text-enhanced cognitive tasks (e.g., reasoning, QA), our TSQA dataset supports both, making it the most comprehensive benchmark. It also covers the widest range of application domains (12 vs. a maximum of 9) and is significantly larger (\textasciitilde200k instances vs. \textasciitilde10k–16k). Furthermore, TSQA includes both real and synthetic data, ensuring greater diversity and robustness for text-enhanced time series analysis.

\begin{table}[!t]
    \centering
    \resizebox{1\linewidth}{!}{
    \fontsize{11}{14}\selectfont
    \setlength{\heavyrulewidth}{1.25pt} 
    \renewcommand{\arraystretch}{1}{
    \begin{tabular}{l l}
        \toprule
        \textbf{Hyperparameter} & \textbf{Assignment} \\
        \midrule
        Base model & Mistral 7B \\
        Computing infrastructure & 1*A100-80GB GPU \\
        Max steps & 4000 \\
        Warm-up steps & 1000 \\
        Batch size per device & 4 \\
        Gradient accumulation steps & 8 \\
        Learning rate & 5e-5 \\
        Embedding learning rate & 1e-5 \\
        Optimizer & AdamW (8-bit) \\
        Learning rate scheduler & Cosine \\
        Weight decay & 0.1 \\ \midrule
        LoRA rank (r) & 16 \\
        LoRA alpha & 16 \\
        LoRA dropout & 0.0 \\
        LoRA target modules & q\_proj, k\_proj, v\_proj, o\_proj, \\
        & gate\_proj, up\_proj, down\_proj \\ \midrule
        Training time & \textasciitilde1 Day \\ 
        \bottomrule
    \end{tabular}}}
    \caption{Hyper-parameters and training time of fine-tuning the Mistral 7B based on the TSQA dataset.}
    \vspace{-5mm}
    \label{tab:lora_hyperparams}
\end{table}

\section{Experiment and Result}

\subsection{Experimental Settings}
In the experiments, we considered the ratio of application domains by randomly selecting 1,400 QA pairs for each task type - forecasting, imputation, anomaly detection, and classification - and by manually reviewing and selecting 1,400 QA pairs for open-ended reasoning, leading to a total of 7,000 QA pairs. We then followed the setting from \cite{cheng2024instruction}. Specifically, to ensure the model adequately learns the time-series domain, we then combined our dataset with a general QA corpus sourced from OpenOrca \cite{OpenOrca} at a 70\% to 30\% ratio, resulting in 10k QA pairs overall. Finally, we formatted all QA pairs so that the question and answer were clearly labeled, and then we tokenized the text. An example of a formatted QA pair from the pre-tokenized text is provided in the Appendix. All training runs were conducted on a single A100 80GB GPU. Table~\ref{tab:lora_hyperparams} shows an overview of the hyperparameters used and training time, using Mistral 7B as an example.

\begin{table*}[!t]
    \centering
    \caption{Comparison of our three fine-tuned models, GPT-4o, and Doubao across diverse tasks. Forecasting and imputation tasks were evaluated using average MSE, while anomaly detection, classification, and open-ended reasoning tasks (including multiple-choice questions (MCQs) and true-false questions (Judgment)) were measured by accuracy. A lower value of MSE $\downarrow$ and a higher value of accuracy $\uparrow$ indicate better performance. $^*$ Doubao uses simple mean forecasting, which outputs the same value for all forecasting.}
    \begin{adjustbox}{width=1\textwidth}
    \fontsize{11}{14}\selectfont
    \renewcommand{\arraystretch}{1}
    \setlength{\heavyrulewidth}{1.25pt} 
    \begin{tabular}{l|cccc|cc}
        \toprule
        \multirow{2}{*}{\textbf{Backbone}} & \multicolumn{4}{c|}{\textbf{Classical Numerical Task}} & \multicolumn{2}{c}{\textbf{Open-Ended Reasoning   QA}} \\ \cline{2-7} 
         & \textbf{Forecasting $\downarrow$} & \textbf{Imputation $\downarrow$} & \textbf{Anomaly Detection $\uparrow$} & \textbf{Classification $\uparrow$} & \textbf{Judgment $\uparrow$} & \textbf{MCQ $\uparrow$} \\ \midrule
        Doubao & ---$^*$ & 0.018 & 0.52 & 0.44 & 0.78 & 0.56 \\
        GPT-4o & 1.79 & 0.018 & 0.64 & 0.32 & 0.72 & 0.58 \\ \midrule
        Llama-3 8B & 2.01 & 0.020 & 0.54 & 0.24 & 0.74 & 0.48 \\
        Qwen-2.5 7B & 1.82 & 0.016 & \textbf{0.68} & \textbf{0.52} & \textbf{0.82} & 0.54 \\
        Mistral 7B & \textbf{1.35} & \textbf{0.014} & 0.58 & 0.44 & 0.80 & \textbf{0.64} \\
        \bottomrule
    \end{tabular}
    \end{adjustbox}
    \label{tab:2}
\end{table*}

\subsection{Results}
\paragraph{Main Results.}
Table \ref{tab:2} presents the performance of three fine-tuned models, alongside the GPT-4o \cite{achiam2023gpt} model and Doubao \cite{volcengine_doubao} results, across multiple tasks. For evaluation, we randomly selected 50 QA pairs for each task type (or question format). 

Overall, our fine-tuned models demonstrated improved performance across multiple tasks. For open-ended reasoning QA, the fine-tuned Qwen model achieved an accuracy of 82\% on judgment questions, while the Mistral model reached 64\% on multiple-choice questions. In forecasting tasks, the MSE values were relatively high, likely due to the long time series in our dataset, which can be challenging for models to handle. Nevertheless, our fine-tuned Mistral model outperformed GPT-4o.

For imputation tasks, the MSE remained low, possibly because having more extensive time series data provides additional information that supports better imputation. Our fine-tuned Mistral model generally showed stronger performance. These results suggest that by continuously pre-training on QA time series pairs, models can learn and leverage time series patterns more effectively.

\begin{figure}[!t]
\begin{center}
\includegraphics[width = 1\linewidth]{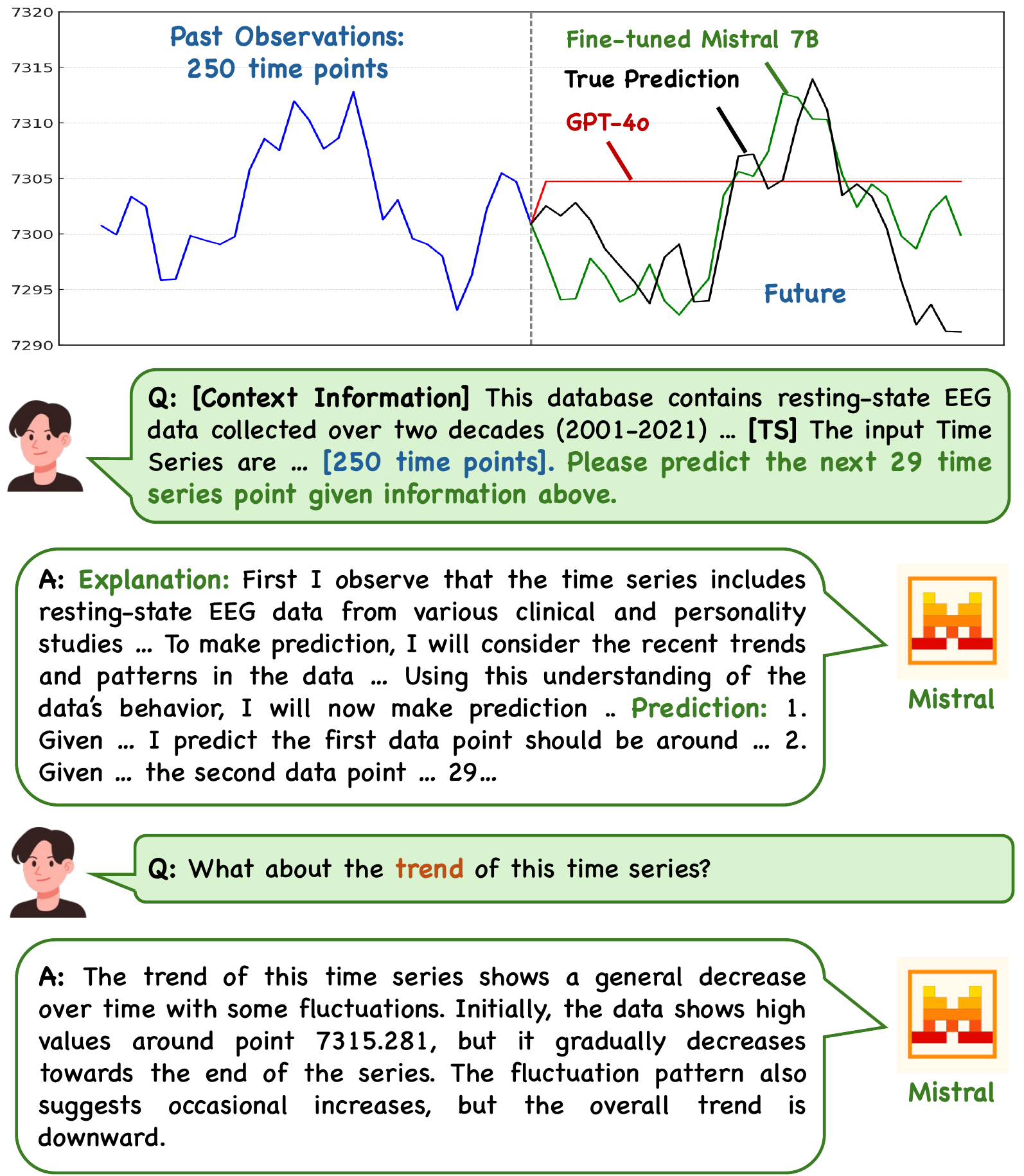}
\end{center}
\caption{Example of long time series forecasting.} \vspace{-1mm}
\label{Figure_5.2_1}
\end{figure}

\paragraph{Forecasting with Long Time Series.}
Figure \ref{Figure_5.2_1} illustrates an example of forecasting results produced by our fine-tuned Mistral 7B model and GPT-4o on a long time-series input. Notably, the Mistral 7B model’s forecasts outperform those generated by GPT-4o. Moreover, the fine-tuned model can provide the rationale behind its predictions. When we posed follow-up questions in natural language, the model offered explanations of its reasoning. This illustrates that the model trained using the TSQA dataset can effectively learn complex time-series patterns and seasonal trends.

\begin{table}[!t]
\centering
\caption{Performance of TSQA-tuned and Zero-shot Mistral 7B on two tasks (evaluated at accuracy).}
\resizebox{0.72\linewidth}{!}{
\fontsize{11}{14}\selectfont
\setlength{\heavyrulewidth}{1.25pt} 
\renewcommand{\arraystretch}{1}{
\begin{tabular}{lcc}
\toprule   
\multicolumn{1}{l}{\textbf{Model}} & \multicolumn{1}{c}{\textbf{Judgment $\uparrow$}} & \multicolumn{1}{c}{\textbf{MCQ $\uparrow$}} \\ \midrule
Zero-shot & 0.78 & 0.60 \\
TSQA-tuned & \textbf{0.80} & \textbf{0.64} \\ \bottomrule
\end{tabular}}} \vspace{-3mm}
\label{tab:mistral_performance}
\end{table}

\paragraph{TSQA-Tuned vs. Zero-Shot Model.} 
To further demonstrate the effectiveness of our fine-tuned model on the proposed TSQA dataset, we compared the performance of the tuned Mistral 7B with that of the original pretrained model on open-ended reasoning QA tasks (including judgment and multiple-choice questions). The results, shown in Table \ref{tab:mistral_performance}, indicate that the TSQA-tuned model performs slightly better than the zero-shot model, suggesting the value of our TSQA dataset for improving question-answering performance.

\begin{figure}[!t]
\begin{center}
\includegraphics[width = 1\linewidth]{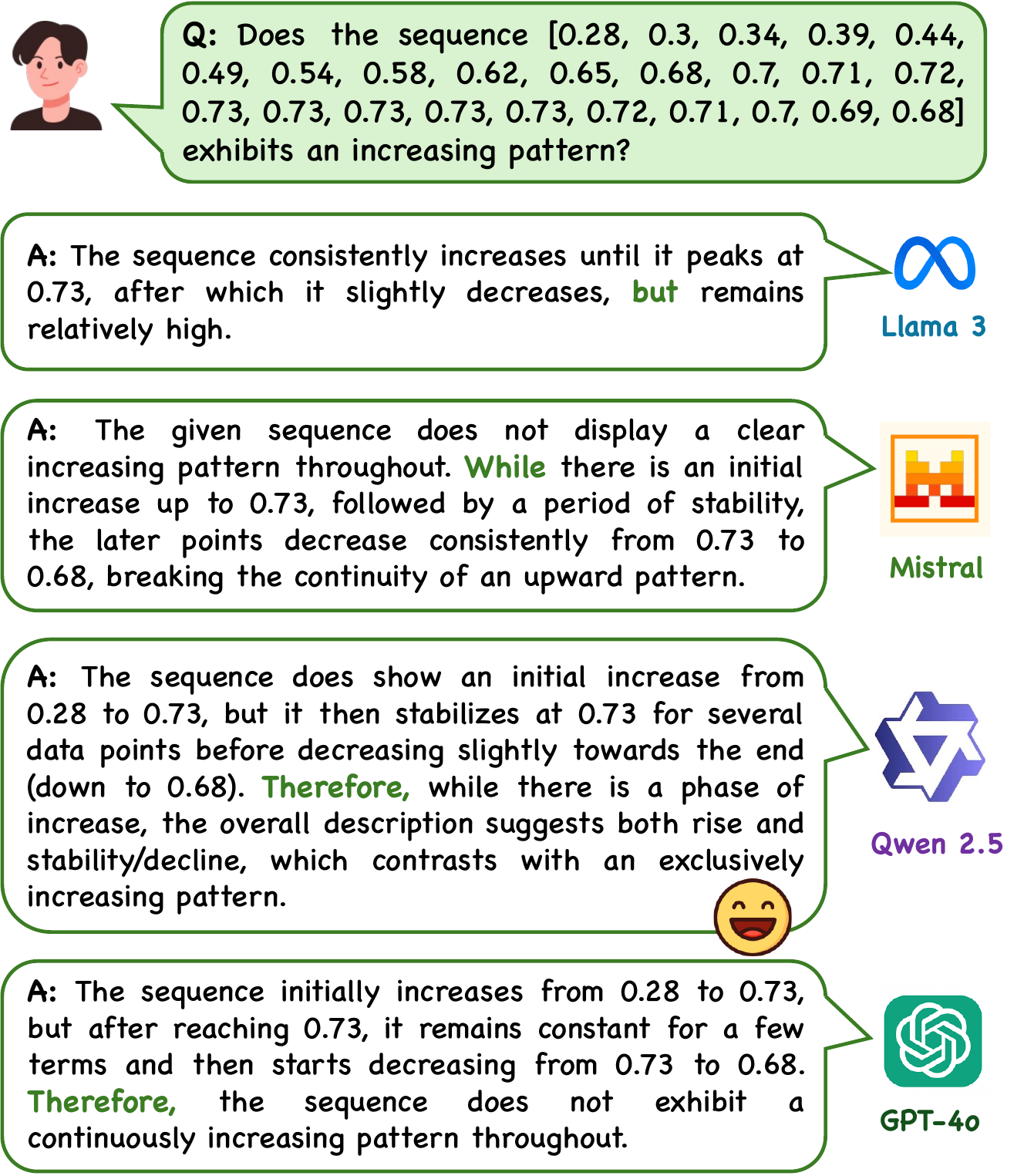}
\end{center} 
\caption{Example of open-ended reasoning QA.}
\label{Figure_5.2_2} \vspace{-3mm}
\end{figure}

\paragraph{Open-Ended Reasoning QA.}
Figure \ref{Figure_5.2_2} presents an example of an open-ended reasoning question answering and compares the responses generated by our fine-tuned models with those from GPT-4o. As shown, our fine-tuned models effectively justify the reasons behind their answers. Notably, the Qwen-2.5 7B model provides a more thorough and comprehensive reasoning process compared to the other models.

\subsection{User Study on Time Series QA}
Given the subjectivity of time series question answering and reasoning tasks, we conducted a user study. We recruited 78 participants from diverse domains, such as data science researchers, financial practitioners, and students of liberal arts, who participated in this study anonymously. The four optional models are A) Llama-3 8B, B) Mistral 7B, C) Qwen-2.5 7B, and D) GPT-4o sequentially. Participants assessed responses based on accuracy, completeness, clarity, and user preference. The complete result and questionnaire of the user study can be found in the Appendix, and structured summaries of the findings are as follows:

\paragraph{Accuracy and Completeness.} Mistral achieved the highest accuracy in tasks requiring numerical precision. For cyclical pattern identification (Q4), it received 80.8\% user preference, and 69.2\% selected its response as most comprehensive for volatility analysis (Q6). However, in trend analysis (Q12), 51.3\% preferred Qwen over Mistral (37.2\%). In summary, Mistral's and Qwen's answers to the numerical questions are generally considered to be more accurate as well as more comprehensive.

\paragraph{Explanation Clarity.} Explanation clarity was measured by asking participants to rate the understandability of the models’ reasoning. In Question 7 (“Considering the data points [60.0, 28.0, …], do you see any seasonal patterns?”), more participants found Mistral’s explanation clear, praising its logical structure and ease of understanding. Although Qwen’s responses were also well received, they trailed slightly behind in these clarity ratings.

\paragraph{User Preference.} In terms of overall user preference, when asked which answer they preferred (as in Question 11), mistral was selected by 70.5\% of participants, while Qwen received 32.1\% of the votes. This reflects a general tendency among users to favor responses that combine detailed data analysis with clear, logical explanations. In contrast to the above two models, Llama and GPT-4o are slightly less well preferred.

Overall, from the user study, Qwen and Mistral emerged as the top models for accuracy-driven tasks, while Mistral excelled in generating thorough explanations and more preferences. This highlights a trade-off between precision and interpretability in open-ended time series reasoning.

\section{Conclusion}
In this paper, we introduce \textit{Time-MQA}, a multi-task framework that unifies time series analysis through natural language question answering, overcoming the limitations of single-task approaches. By integrating forecasting, imputation, anomaly detection, classification, and open-ended reasoning, \textit{Time-MQA} enables flexible interactions with time-series data. The introduced \textit{TSQA} dataset comprises $\sim$200k question-answer pairs across twelve diverse domains and varying time series lengths. It is a vital resource for advancing time series question answering and reasoning in LLMs. Experimental results show that continually pre-training models such as Mistral 7B, Llama 8B, and Qwen-2.5 7B on \textit{TSQA} enhance their ability to interpret temporal patterns — going beyond numeric tasks to generate contextually rich insights. This work bridges the gap between traditional time series analysis and modern LLM capabilities, making temporal information more accessible and context-aware. Future directions include expanding the \textit{TSQA} dataset to dynamic real-world data streams and exploring strategies to further refine time series reasoning.

\clearpage
\section*{Limitations}
While Time-MQA advances multi-task time series question answering, it has limitations. Our proposed TSQA dataset, though diverse, may not cover all real-world scenarios, particularly highly irregular or domain-specific time series. But data with those properties may be common in real scenarios~\cite{mulayim2024time}. Our financial datasets are currently limited to earnings call transcripts. Future research could enrich the financial dataset by incorporating other types of financial time series and textual data, such as Jerome Powell's speeches and their impact on the S\&P 500. Our approach relies on pre-training LLM architectures, which may not be optimized for long numerical sequences, which may cause hallucination problems~\cite{yao2023llm}. Evaluating open-ended reasoning remains challenging, as current automated metrics may not fully capture response quality~\cite{kong2025position}. Additionally, computational costs for fine-tuning large models are high, which may limit some real-time applications~\cite{jin2024position}.

\section*{Ethics Statement}
All datasets and language models used in this work are publicly available. The TSQA dataset was constructed from well-established, publicly accessible time series benchmarks and synthetic data generation followed ethical guidelines to minimize biases and ensure data quality. 

\section*{Acknowledgments}
We would like to thank all reviewers for their insightful comments and suggestions to help improve the paper. Y.K. gratefully acknowledges the computational support provided by the Biomedical Research Computing (BMRC) and the Integrative Computational Biology Group at the Centre for Human Genetics, University of Oxford.

\bibliography{acl_latex}

\newpage

\appendix

\section{Examples of TSQA Dataset}
\label{sec:appendix}
Below are several examples from the TSQA dataset. 

\subsection{Forecasting.}
\textbf{Question:} The Self-regulation of Slow Cortical Potentials dataset, provided by the University of Tuebingen, involves EEG recordings from a healthy subject controlling a cursor on a screen through cortical potentials. The sampling frequency of this dataset is 256 Hz. The input Time Series are \textbf{\textit{[Time Series Data Points]}}. Please predict the next nine time series points given the information above. 

\textbf{Answer:} Based on the given information, the predictions are [-11.69, -10.72, -9.84, -9.88, -9.03, -8.78, -8.41, -8.28, -7.59].
\\~\\
\textbf{Question:} From March 31, 2014 to June 04, 2014, the weighted ILI rate was 1.74542, and the ili total count reached 14637 across 1866 providers serving a total of 797863 patients, especially 0-4 age group was 3582, 5-24 age group was 5659, 25-49 age group was 3153, 50-64 age group was 1436, 65+ age group was 807, and 25-64 age group was unavailable. The internet search results for the same period are summarized as follows: ... The prior historical average weighted ILI rate during this period was 1.286616667 with a standard deviation of 0.073690536. Please analyze the data and provide a target result on June 04, 2014 with a summary of the findings.

\textbf{Answer:} Based on the provided textual information, I predict that the influenza infection situation in the United States for the upcoming period is likely to be moderate to severe, with a potential for increased risk of complications and co-infections, particularly among vulnerable populations such as those with cystic fibrosis and critically ill patients. Therefore, the predicted weighted ILI rate for June 04, 2014 is estimated to be 1.83453.

\subsection{Imputation.}
\textbf{Question:} This dataset, sourced from the Kaggle web traffic forecasting competition, contains daily time series data representing the number of hits for 145,063 Wikipedia pages from July 1, 2015, to September 10, 2017. The input Time Series are \textbf{\textit{[Time Series Data Points with Missing Values]}}. Please give full-time series with missing value imputed. The missing values are represented by 'X'.

\textbf{Answer:} Based on the given information, the full-time series with missing value imputed are \textbf{\textit{[Time Series Data Points]}}.

\subsection{Classification.} 
\textbf{Question:} Following data provides accelerometer data for activity recognition research. The dataset has a sampling rate of 20Hz and records accelerometer data for six activity states: walking, jogging, sitting, standing, upstairs, and downstairs. Each sample includes acceleration values for the X, Y, and Z axes, ranging from -20 to 20, where 10 represents 1g and 0 indicates no acceleration. The recorded acceleration includes gravitational acceleration, so when the phone is stationary on a flat surface, the vertical axis registers approximately ¬±10. We provide 10 timestamps of accelerometer data, with each timestamp containing X, Y, and Z values, for a total of 30 values. The recorded Time Series is \textbf{\textit{[Time Series Data Points]}}. Please judge whether this data corresponds to 'Walking' or 'Jogging' or 'Upstairs' or 'Downstairs' or 'Sitting' or 'Standing' given the information above.

\textbf{Answer:} Based on the given information, the activity is Walking.

\subsection{Anomaly Detection.}
\textbf{Question:} The following data is derived from AWS CloudWatch services and is primarily used for monitoring and managing metrics related to cloud infrastructure and application performance. These include CPU usage, memory consumption, network throughput, and disk I/O metrics. This data can be utilized to identify performance issues in cloud services, such as traffic spikes, resource exhaustion, or network anomalies, and to perform root cause analysis for rapid problem resolution. The input Time Series is \textbf{\textit{[Time Series Data Points]}}. Please determine whether there are anomalies in this time series given the information above.

\textbf{Answer:} Based on the given information, this time series includes Anomaly Points.

\subsection{Open-Ended Reasoning QA.}
\textbf{Question:} Summarize the overall behavior observed in the data points \textbf{\textit{[Time Series Data Points]}}. 

\textbf{Answer:} The data initially shows a slow increase followed by a decreasing trend. The middle part of the data points represents a peak with few higher fluctuations and later shows a slightly decline indicating a volatile pattern.
\\~\\
\textbf{Question:} Determine if there's a structural break in the data points \textbf{\textit{[Time Series Data Points]}}. A) Yes, between 0.71 and 0.72 B) Yes, between 0.65 and 0.63 C) No structural break

\textbf{Answer:} C) No structural break. The data follows a gradual upward trend and subsequent flattening with slight variability, but no abrupt change in the level or direction indicating a structural break.
\\~\\
\textbf{Question:} The data points \textbf{\textit{[Time Series Data Points]}} exhibit a clear seasonal pattern. True or False?

\textbf{Answer:} False. While there are fluctuations and some repeated values, no clear cyclical pattern emerges that would indicate seasonality. The variations appear to be more random than driven by any seasonal influences.


\section{Open-Ended Reasoning QA Generation}
\begin{figure}[ht]
\begin{center}
\includegraphics[width = 1\linewidth]{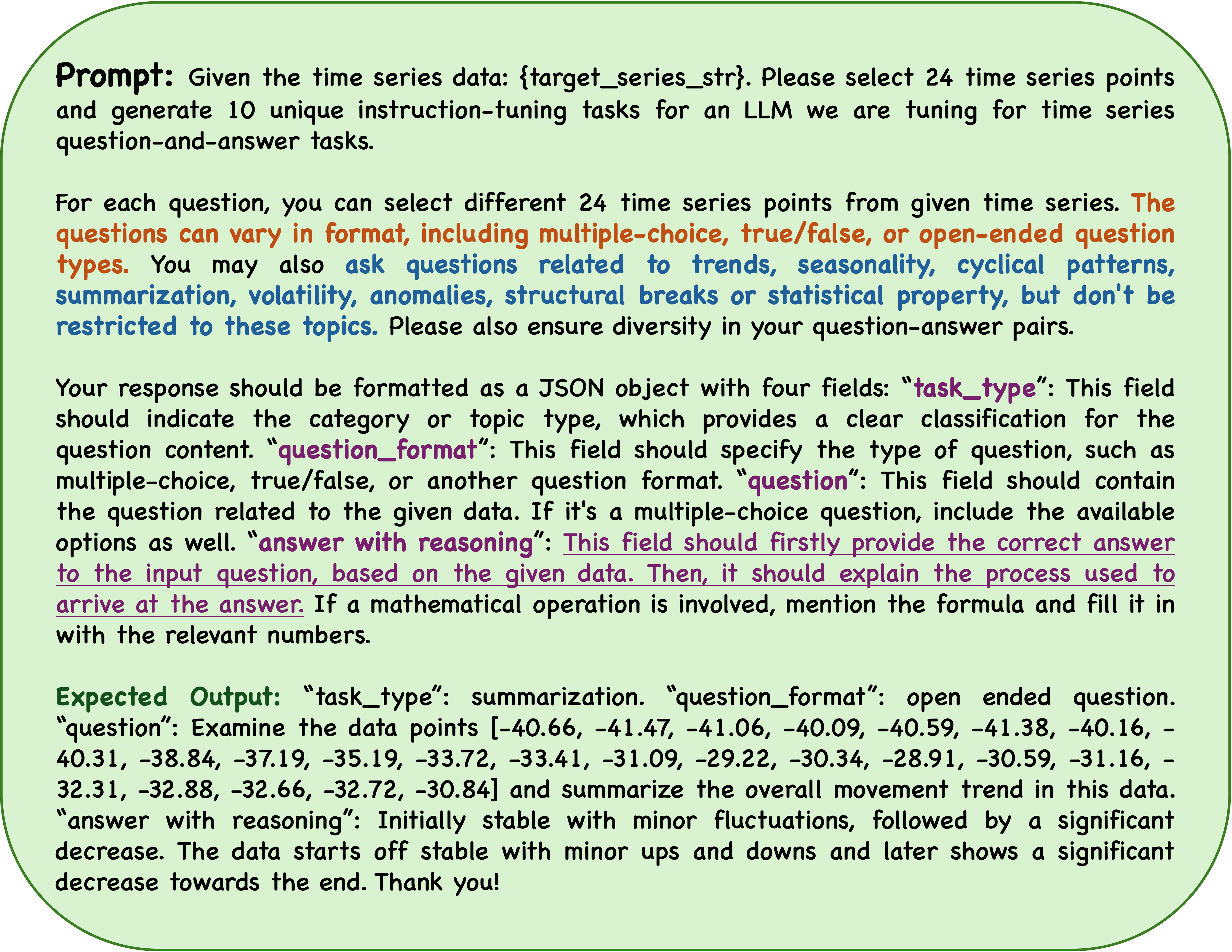}
\end{center}
\caption{Prompt for generating open-ended reasoning QA pairs.}
\label{Figure_appendix_1}
\end{figure}

Open-ended reasoning QA pairs were partially generated by GPT-4o to broaden the range of question types. However, the time-series signals themselves are derived from real measurements. Figure \ref{Figure_appendix_1} illustrates the prompt used to generate open-ended reasoning QA pairs.

\section{High-Level Data Statistics for the TSQA dataset}
Table \ref{figure_appendix_2} presents high-level data statistics for the TSQA dataset, including the number of QA pairs, distribution across tasks and domains, the type of real versus synthetic data, and various data sources. These sources include UTSD \cite{liu2024timer}, TimeMMD \cite{liutime}, WISDM \cite{kwapisz2011activity}, FOG \cite{bachlin2009potentials}, NAB \cite{ahmad2017unsupervised}, MIT-BIH Arrhythmia (ECG) \cite{moody2001impact}, UCR \cite{dau2018yanping}, MGAB \cite{thill2020markusthill}, SensorScope \cite{barrenetxea2019sensorscope}, KPI \cite{aiops2018kpi}, and Yahoo \cite{yahoo2021s5}.

\begin{table}[ht]
\centering
\caption{High-level data statistics of the TSQA dataset.}
\resizebox{1\linewidth}{!}{
\fontsize{10}{13}\selectfont
\setlength{\heavyrulewidth}{1.25pt}
\renewcommand{\arraystretch}{1.15}{
\begin{tabular}{llcll}
\toprule
\textbf{Task} & \textbf{Domain} & \textbf{Real vs Synth} & \textbf{Count} & \textbf{Data Source} \\
\midrule
\multirow{9}{*}{Forecasting} 
    & Healthcare   & Real           & 19980     & UTSD, TimeMMD \\
    & Web         & Real           & 11362                  & UTSD \\
    & Nature      & Real           & 7038                   & UTSD \\
    & Energy      & Real           & 1545         & UTSD, TimeMMD \\
    & Transport   & Real           & 62                     & UTSD \\
    & IoT         & Real           & 36                     & UTSD \\
    & Environment & Real           & 18                     & UTSD \\
    & Finance     & Real           & 1987                   & TSQA (Our) \\
    & Traffic     & Real           & 529                    & TimeMMD \\
\midrule
\multirow{7}{*}{Imputation}
    & Healthcare   & Real           & 18800                  & UTSD \\
    & Web         & Real           & 12397                  & UTSD \\
    & Nature      & Real           & 7275                   & UTSD \\
    & Energy      & Real           & 65                     & UTSD \\
    & Transport   & Real           & 65                     & UTSD \\
    & IoT         & Real           & 36                     & UTSD \\
    & Environment & Real           & 19                     & UTSD \\
\midrule
Classification & Human Activities & Real & 37000 & WISDM, FOG \\
\midrule
\multirow{7}{*}{Anomaly Detection}
    & AIOps           & Real       & 10161                  & NAB-realAWSCloudwatch, KPI\\
    & Web             & Real       & 3349                   & Yahoo \\
    & Environment     & Real       & 12547                  & Sensorscope \\
    & Finance         & Real       & 43                     & NAB-realAdExchange \\
    & Healthcare      & Real       & 10057                  & MIT-BIH Arrhythmia (ECG) \\
    & Machine Sensors & Real/Synth & 721                    & UCR, MGAB \\
    & Traffic         & Real       & 122                    & NAB-realTraffic \\
\midrule
\multirow{7}{*}{Open-Ended QA}
    & Energy        & Real\&Synth         & 5000   & UTSD, TSQA (Our) \\
    & Transport     & Real\&Synth         & 5000   & UTSD, TSQA (Our) \\
    & Nature        & Real\&Synth         & 5000   & UTSD, TSQA (Our) \\
    & Web           & Real\&Synth         & 5000   & UTSD, TSQA (Our) \\
    & IoT           & Real\&Synth         & 5000   & UTSD, TSQA (Our) \\
    & Environment   & Real\&Synth         & 5000   & UTSD, TSQA (Our) \\
    & Healthcare    & Real\&Synth         & 7629   & UTSD, TSQA (Our) \\
\bottomrule
\end{tabular}}}
\label{figure_appendix_2}
\end{table}

\section{Training Data Format}

We format our question-and-answer pairs using a specifically designed template to clearly separate questions from answers. The template is structured as follows: \texttt{<QUE> \{Question\} <ANS> \{Answer\} </END>}. Additionally, for continual pre-training of the Llama model, we prepend \texttt{<|begin\_of\_text|>} at the start of each sample and append \texttt{<|end\_of\_text|>} at the end. For the Mistral model, we use \texttt{<s>} at the beginning of each sample and \texttt{</s>} at the end. In the case of the Qwen model, only \texttt{<|endoftext|>} is added at the end of each sample. 

\vspace{3mm}

\section{Experiments on Model Reliance on Time Series Data}
To verify the model's reliance on time-series data, we conducted a set of ablation experiments. Specifically, we selected 50 Multiple Choice Questions (MCQs) and 50 True-or-False questions that targeted key time-series characteristics such as seasonality, abrupt changes, trends, and anomalies. We evaluated both our fine-tuned Mistral 7B and Qwen-2.5 7B models under the following experimental setups:
\\~\\
\textbf{Exp A. Default (Time Series + Questions):} The original setup containing both the time series and the questions.
\\~\\
\textbf{Exp B. Time Series Removed + Questions:} Only the question text is provided; all numeric data are omitted.
\\~\\
\textbf{Exp C. Random Time Series + Questions:} The original time series is replaced with random Gaussian values, while the question remains unchanged.
\\~\\
\textbf{Exp D. Shuffled Time Series + Questions:} The order of the original time series values is shuffled, destroying temporal structure but retaining the same set of values.
\\~\\
For Experiments C and D, we manually determined “correct” ground-truth labels by inspecting the randomly generated or shuffled series. For instance, if a time series originally exhibited a strong upward trend that disappeared after shuffling, the ground-truth label was set to “no trend.”

Table~\ref{tab:ablation_results} summarizes our findings. ``N/A'' indicates cases where the model outputs were highly inconsistent or typically refused, making it infeasible to compute an accuracy score.

\begin{table}[!h]
\centering
\caption{The results of experiments on model reliance on time series data. The accuracies of Qwen-2.5 7B and Mistral 7B on True/False and MCQ questions are reported under different experimental settings. Higher value of accuracy indicates better performance.}
\resizebox{1\linewidth}{!}{
\fontsize{11}{13}\selectfont
\setlength{\heavyrulewidth}{1.25pt}
\renewcommand{\arraystretch}{1.15}{
\begin{tabular}{llcc}
\toprule
\textbf{Model} & \textbf{Experiment} & \textbf{True/False Accuracy} & \textbf{MCQ Accuracy} \\
\midrule
\multirow{4}{*}{Qwen-2.5 7B}
    & Exp A: Default & 0.82 & 0.54 \\
    & Exp B: TS Removed & N/A & N/A \\
    & Exp C: Random TS & 0.98 & 0.74 \\
    & Exp D: Shuffled TS & 0.84 & 0.58 \\
\midrule
\multirow{4}{*}{Mistral 7B}
    & Exp A: Default & 0.80 & 0.64 \\
    & Exp B: TS Removed & N/A & N/A \\
    & Exp C: Random TS & 0.96 & 0.70 \\
    & Exp D: Shuffled TS & 0.82 & 0.60 \\
\bottomrule
\end{tabular}}}
\label{tab:ablation_results}
\end{table}

In \textbf{Exp B} (Time Series Removed), both models typically produced incomplete or fabricated answers, making it impossible to consistently score outputs. This strongly suggests that, without numeric data, the models cannot reliably answer time-series–specific questions.

In \textbf{Exp C} (Random Time Series), the randomly generated values generally lacked meaningful seasonality or discernible trends. The models appeared to recognize this absence of structure, frequently defaulting to statements such as “no distinct pattern” or “no strong seasonality,” which were often accurate given the lack of cyclical behavior. As a result, higher true-or-false accuracy was sometimes observed — not because of deeper model understanding, but because the unstructured data made generic, correct answers more likely. Notably, the models’ responses shifted when the series was randomized, indicating they were actively parsing the numeric inputs.

Finally, in \textbf{Exp D} (Shuffled Time Series), shuffling effectively removed any evident trend or abrupt break present in the original data. Consequently, the models frequently adjusted their answers. This further supports that the models actively read and interpret the numeric data, rather than simply ignoring it.

In summary, these findings could indicate that our models reflect on the time-series inputs and adapt their responses accordingly when the numeric context is altered or removed.

\vspace{4mm}
\section{User Study}
A total of 78 surveys were collected, and the results for each question are as follows:

\paragraph{Question 1.} 
What is your field of study?

\begin{figure}[ht]
\begin{center}
\includegraphics[width = 1\linewidth]{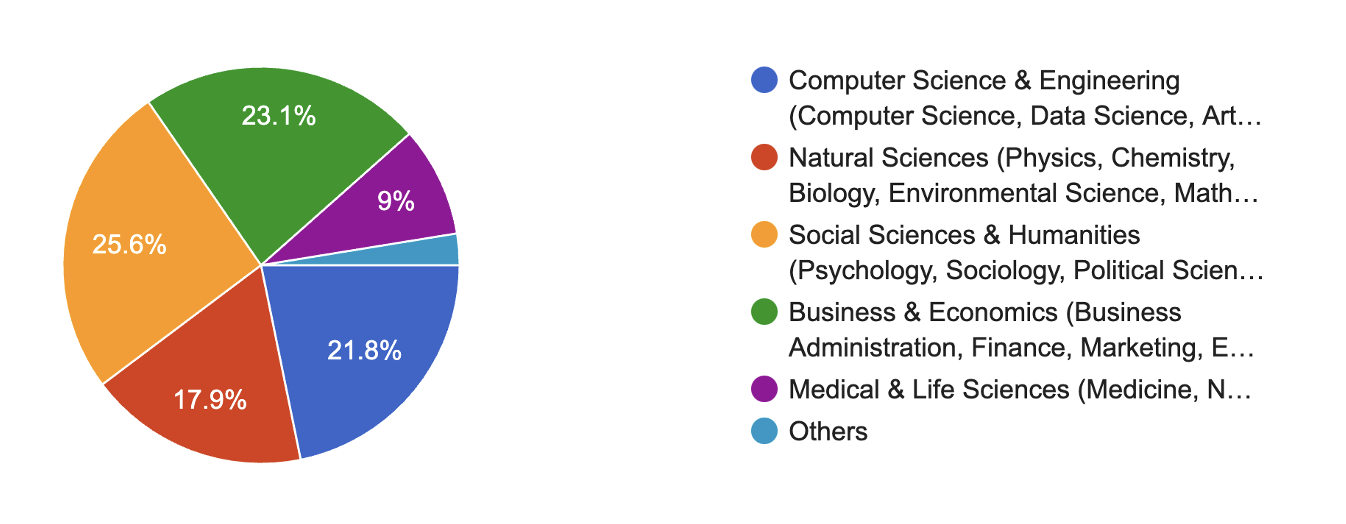}
\end{center}
\label{Figure_q1}
\end{figure}

\paragraph{Question 2.}
How familiar are you with time series analysis? (1 = Not at all, 5 = Very familiar)

\begin{figure}[ht]
\begin{center}
\includegraphics[width = 0.9\linewidth]{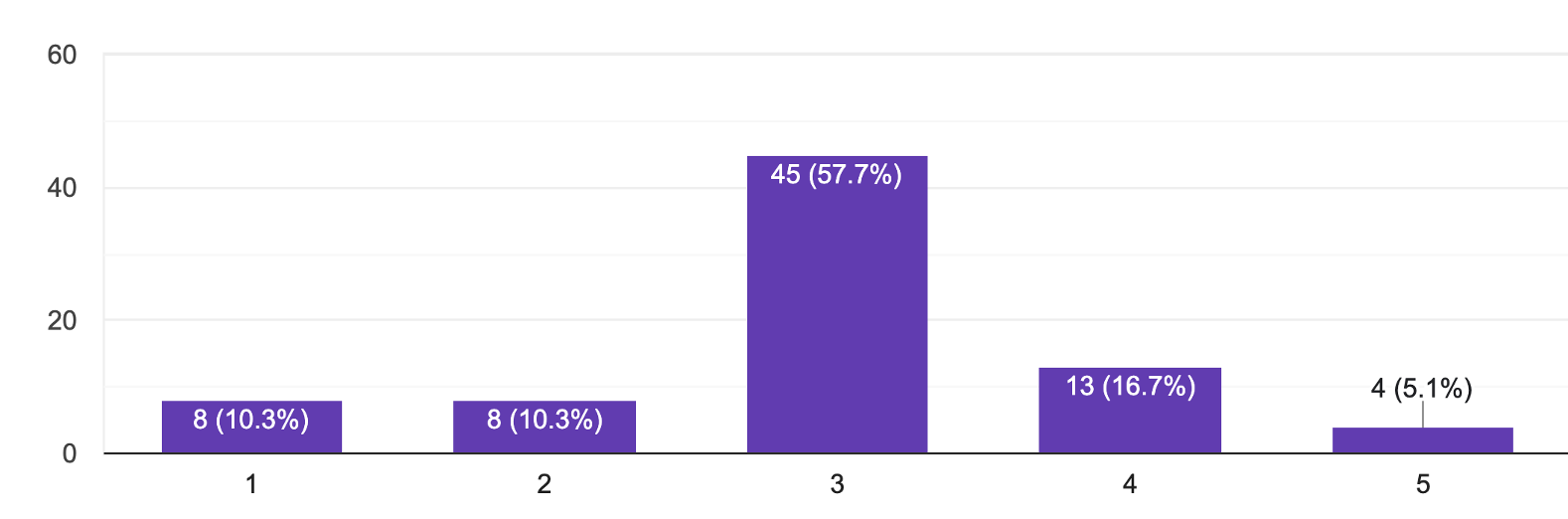}
\end{center}
\label{Figure_q2}
\end{figure}

\paragraph{Question 3.}
How accurate is the answer with respect to the question? 
\\~\\
Analyse and summarise the trend and pattern in the data set [68.0, 83.0, 95.0, 103.5, 108.5, 112.0, 113.0, 114.0, 114.5, 115.5, 116.0, 116.0, 115.0, 114.0, 113.5, 111.5, 108.0, 102.5, 100.5, 93.5, 88.0, 82.0, 78.0, 73.5].

\begin{figure}[ht]
    \centering
    \begin{subfigure}[b]{\linewidth}
        \includegraphics[width=0.9\linewidth]{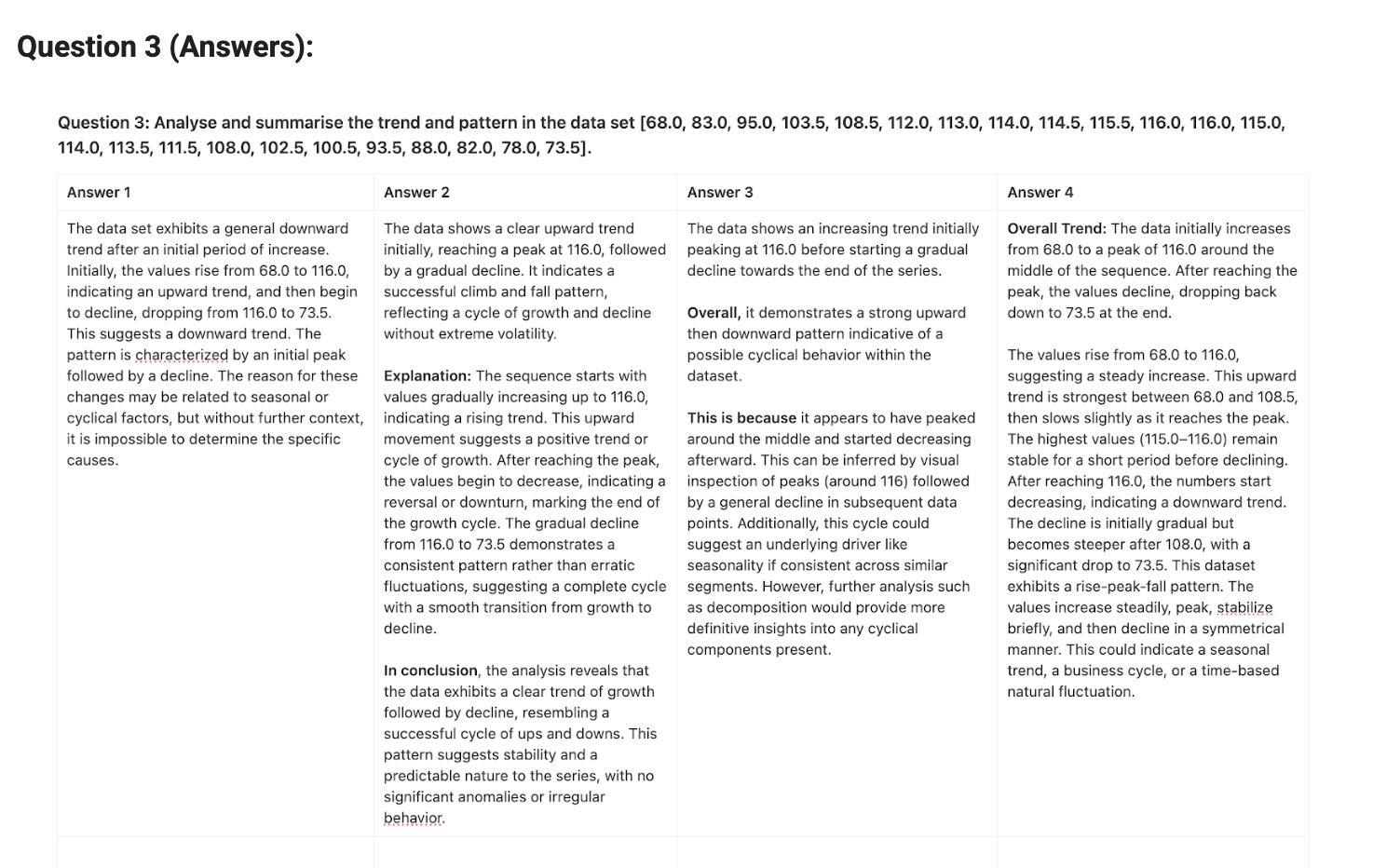}
        \label{fig:q3_1}
    \end{subfigure}
    \vspace{0.5cm} 
    \begin{subfigure}[b]{\linewidth}
        \includegraphics[width=0.9\linewidth]{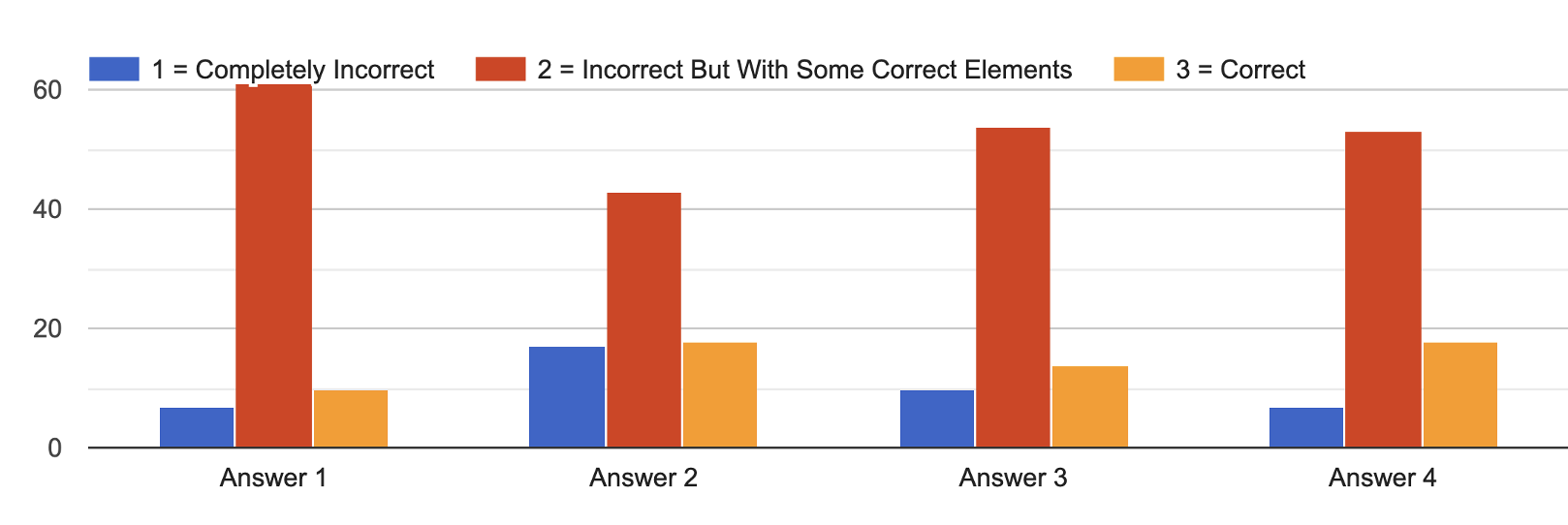}
        \label{fig:q3_2}
    \end{subfigure}
    \label{fig:q3_combined}
    \vspace{-10mm}
\end{figure}

\paragraph{Question 4.}
Which do you think is the most accurate answer with respect to the open-ended reasoning question?
\\~\\
Identify any cyclical patterns in the data points [0.28, 0.27, 0.26, 0.25, 0.24, 0.23, 0.24, 0.25, 0.26, 0.28, 0.3, 0.33, 0.36, 0.39, 0.43, 0.46, 0.5, 0.53, 0.57, 0.6, 0.63, 0.65, 0.67, 0.68].

\begin{figure}[ht]
    \centering
    \begin{subfigure}[b]{\linewidth}
        \includegraphics[width=0.9\linewidth]{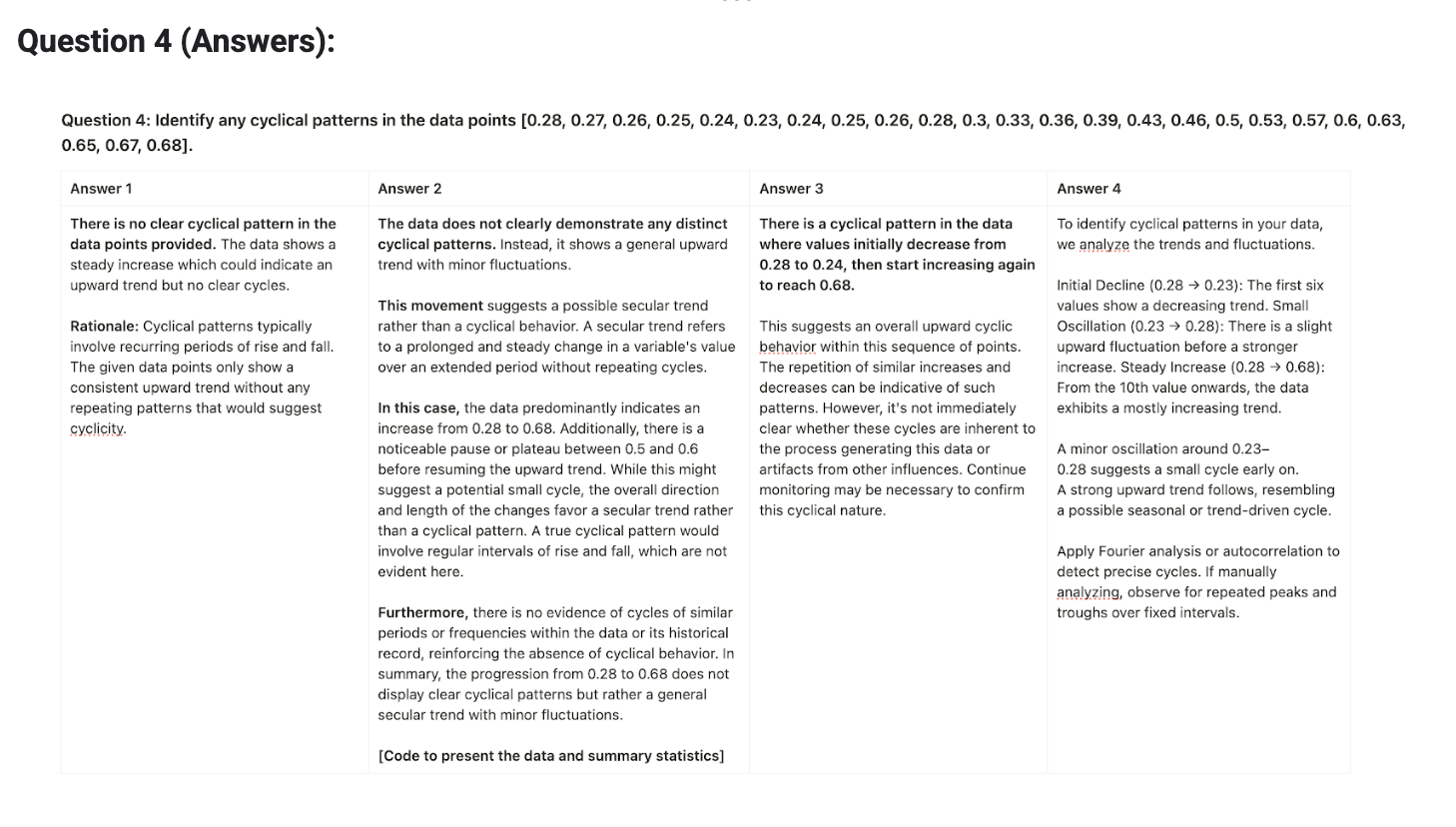}
        \label{fig:q3_1}
    \end{subfigure}
    \vspace{0.5cm} 
    \begin{subfigure}[b]{\linewidth}
        \includegraphics[width=0.9\linewidth]{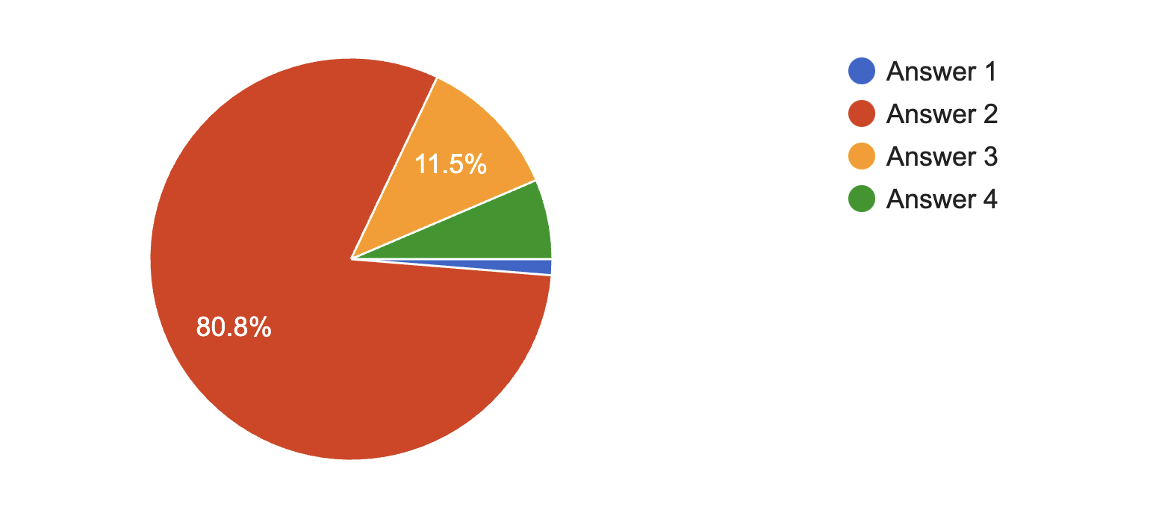}
        \label{fig:q3_2}
    \end{subfigure}
    \label{fig:q3_combined}
    \vspace{-10mm}
\end{figure}

\paragraph{Question 5.}
How complete is the answer? Did it address all parts of the question?
\\~\\
Examine the data points [19.34, 20.41, 20.38, 19.28, 19.75, 22.84, 25.09, 24.97, 24.75, 26.25, 27.0, 25.03, 22.41, 21.78, 22.44, 21.19, 19.44, 19.62, 20.62, 20.16, 18.03, 17.19, 18.31, 19.91] and summarize the overall movement trend in this data.

\begin{figure}[ht]
    \centering
    \begin{subfigure}[b]{\linewidth}
        \includegraphics[width=0.9\linewidth]{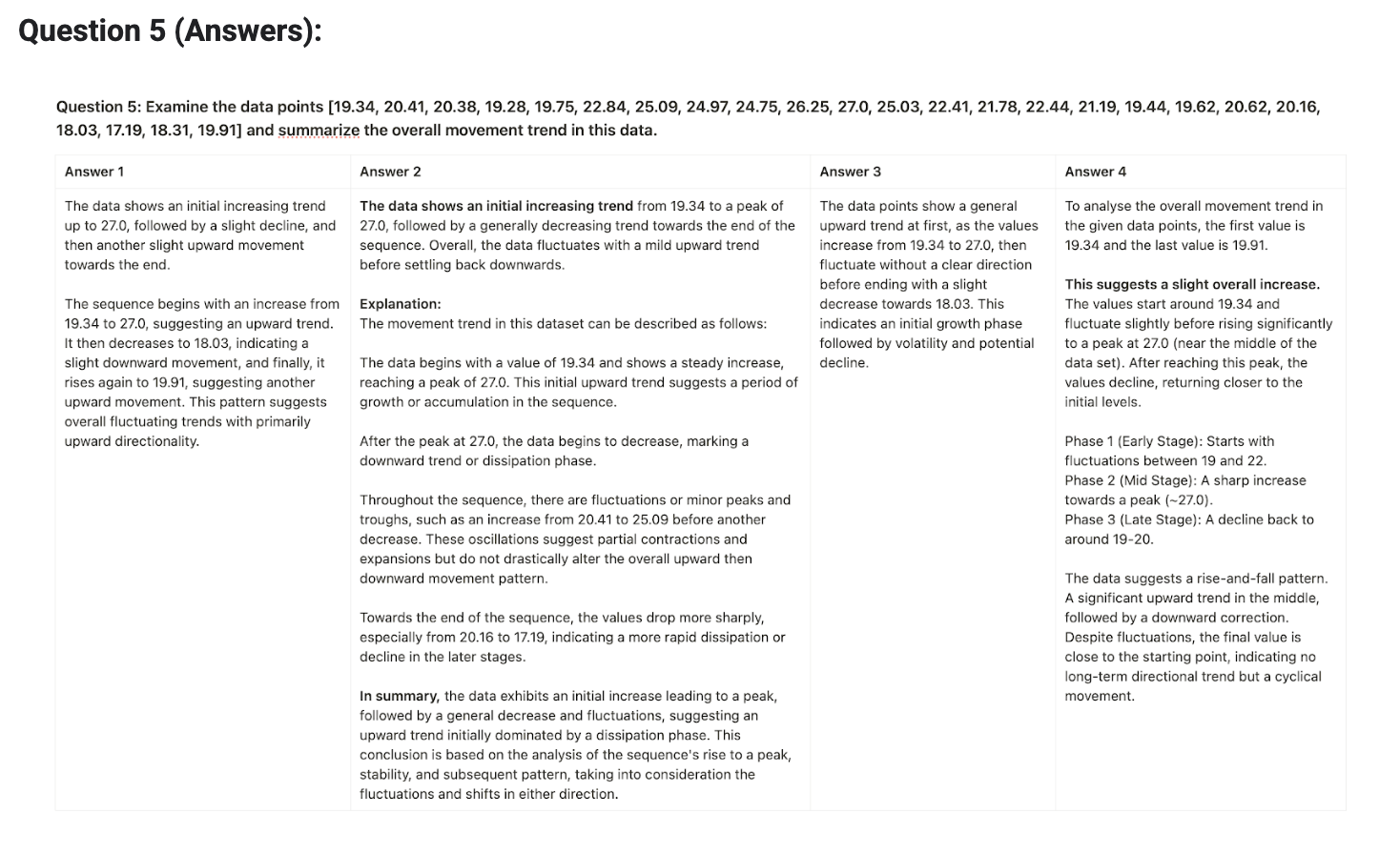}
        \label{fig:q3_1}
    \end{subfigure}
    \vspace{0.5cm} 
    \begin{subfigure}[b]{\linewidth}
        \includegraphics[width=0.9\linewidth]{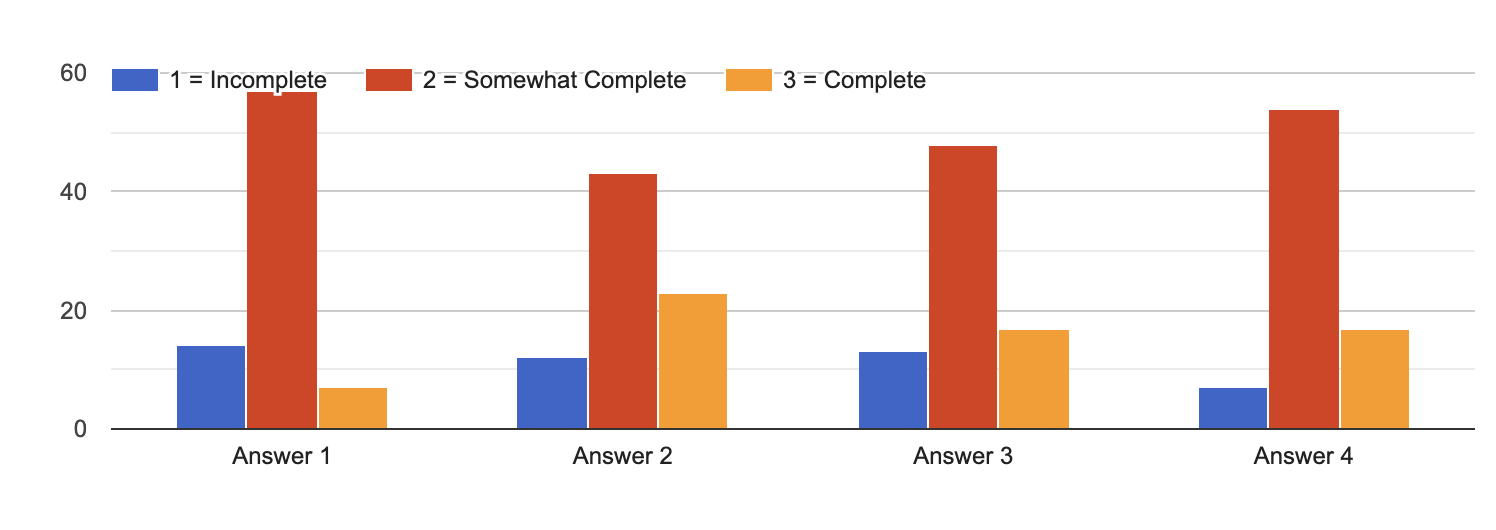}
        \label{fig:q3_2}
    \end{subfigure}
    \label{fig:q3_combined}
    \vspace{-10mm}
\end{figure}

\paragraph{Question 6.}
Which answer is the most comprehensive in relation to the question?
\\~\\
Considering the data points [47.0, 51.0, 51.0, 20.0, 48.0, 44.0, 41.0, 28.0, 48.0, 38.0, 67.0, 66.0, 70.0, 54.0, 63.0, 56.0, 76.0, 58.0, 60.0, 58.0, 69.0, 56.0, 39.0, 50.0], which statement best describes the volatility of this time series?
A) The volatility is low with consistent data points.
B) There is moderate volatility with regular patterns.
C) High volatility with sudden peaks and drops.
D) The time series exhibits a perfectly linear trend.

\begin{figure}[ht]
    \centering
    \begin{subfigure}[b]{\linewidth}
        \includegraphics[width=0.9\linewidth]{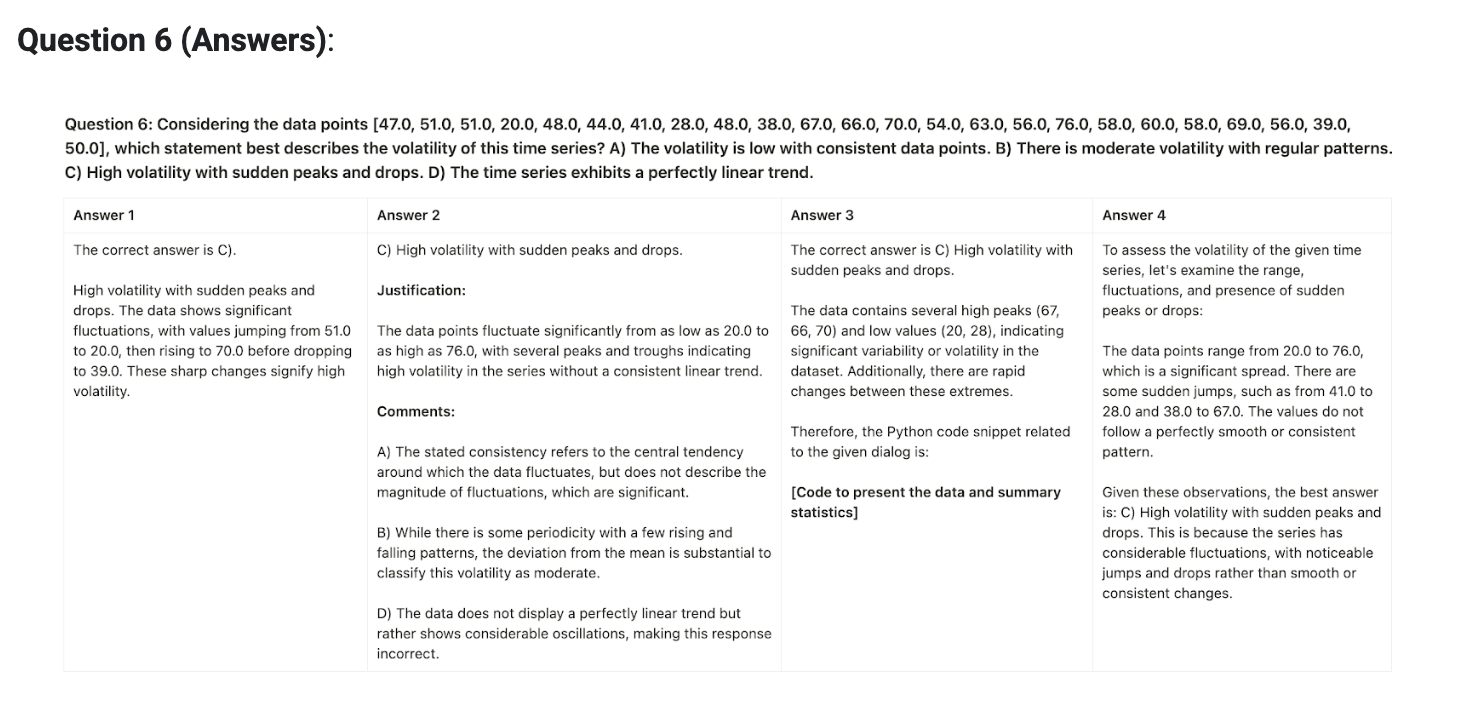}
        \label{fig:q3_1}
    \end{subfigure}
    \vspace{0.5cm} 
    \begin{subfigure}[b]{\linewidth}
        \includegraphics[width=0.9\linewidth]{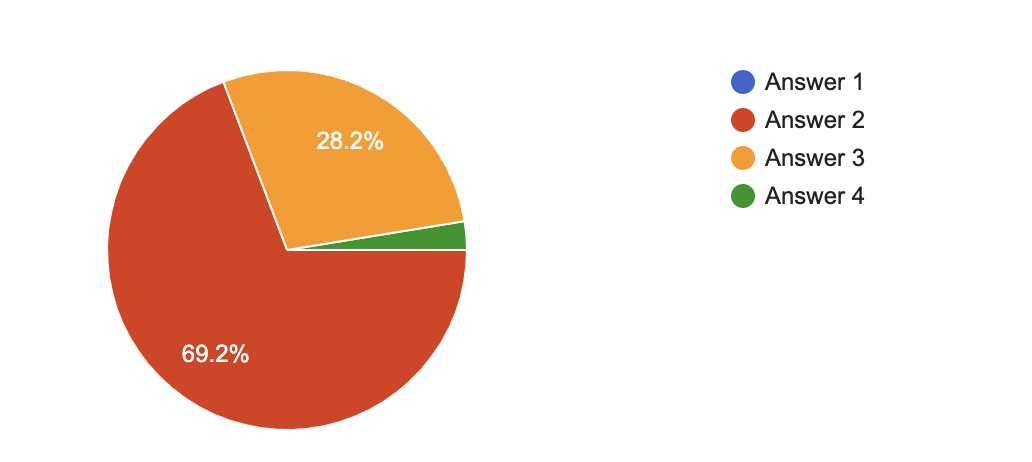}
        \label{fig:q3_2}
    \end{subfigure}
    \label{fig:q3_combined}
\end{figure}

\paragraph{Question 7.}
How clear or understandable is the reasoning provided in the answer?
\\~\\
Considering the data points [60.0, 28.0, 46.0, 51.0, 37.0, 46.0, 41.0, 35.0, 32.0, 66.0, 10.0, 21.0, 67.0, 72.0, 83.0, 61.0, 11.0, 31.0, 62.0, 50.0, 56.0, 76.0, 53.0, 86.0], do you see any seasonal patterns?
A) Yes, there is a seasonal pattern.
B) No, there is no identifiable seasonal pattern.

\begin{figure}[ht]
    \centering
    \begin{subfigure}[b]{\linewidth}
        \includegraphics[width=0.9\linewidth]{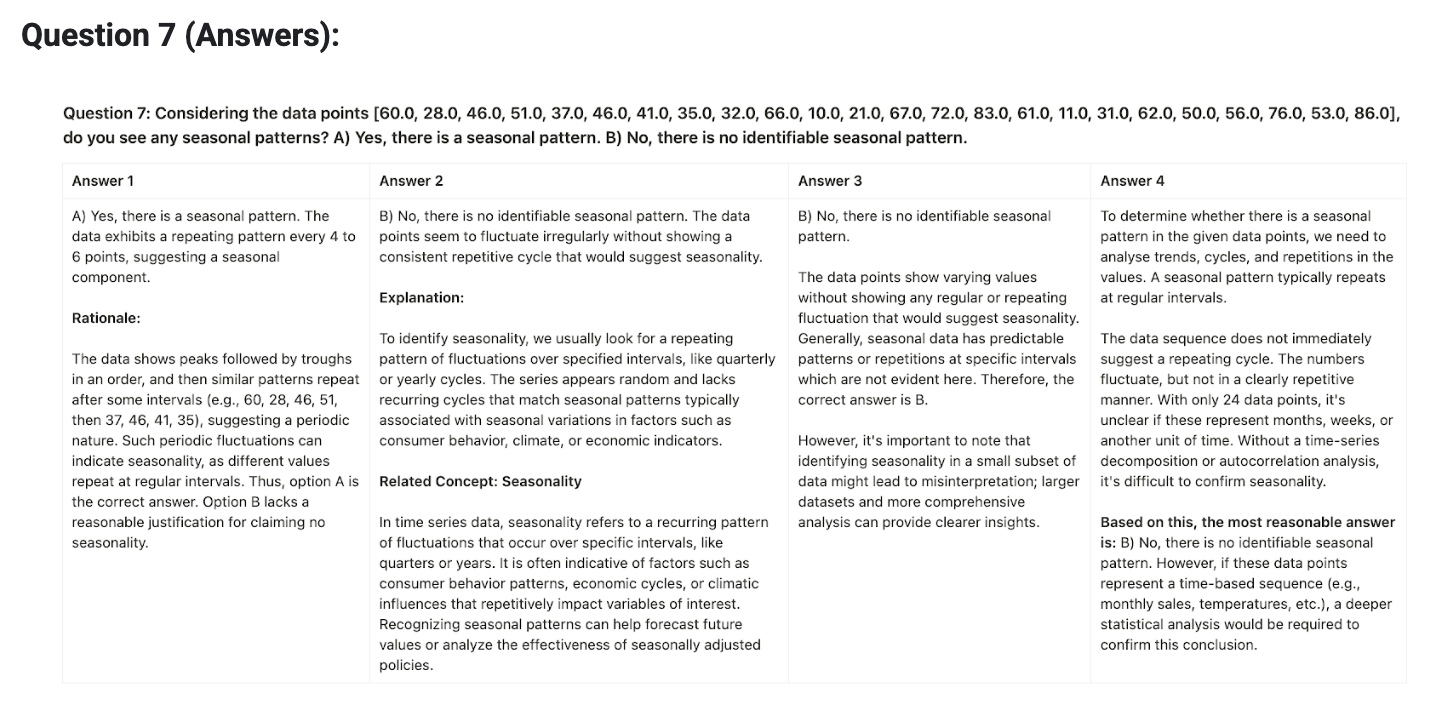}
        \label{fig:q3_1}
    \end{subfigure}
    \vspace{0.5cm} 
    \begin{subfigure}[b]{\linewidth}
        \includegraphics[width=0.9\linewidth]{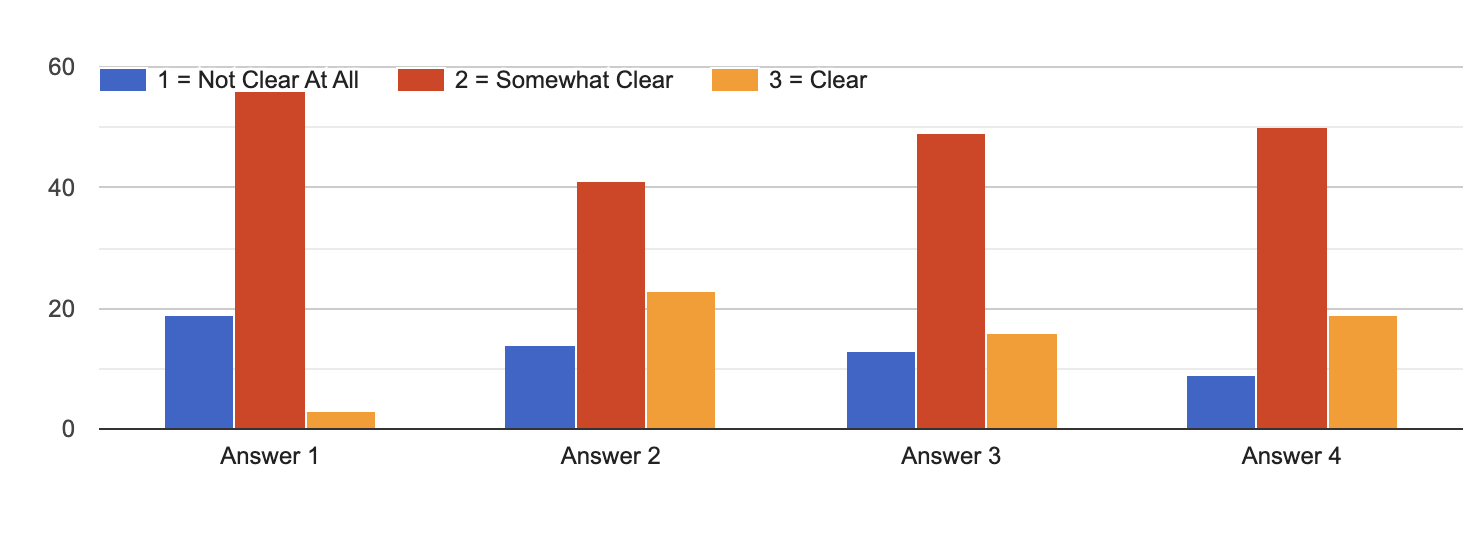}
        \label{fig:q3_2}
    \end{subfigure}
    \label{fig:q3_combined}
    \vspace{-10mm}
\end{figure}

\paragraph{Question 8.}
How clearly does the response explain its reasoning? (1 = Not Clear At All, 5 = Really Clear)
\\~\\
What is the volatility level in the dataset [0.83, 0.8, 0.76, 0.75, 0.75, 0.76, 0.78, 0.8, 0.8, 0.8, 0.81, 0.8, 0.8, 0.8, 0.8, 0.79, 0.8, 0.82, 0.83, 0.82, 0.82, 0.82, 0.82, 0.81]? A) High B) Moderate C) Low D) No volatility"
\\~\\
Answer: C) Low
Explanation: The dataset shows moderate stability with minor fluctuations, indicating a low volatility level. The changes between consecutive points are generally small and consistent rather than erratic or extreme. Thinking process: The dataset has some movement, but it is not highly variable or erratic. There are no significant spikes or drops across the data points. The changes are steady and minor, suggesting a low level of volatility. Comparing to other options:
\\~\\
A) High: The dataset does not show high levels of volatility with major fluctuations.
\\~\\
B) Moderate: While there is some movement, it is not as severe as to classify it as moderate volatility.
\\~\\
D) No volatility: There is some movement, but it is not so minimal that it would be considered no volatility.
\\~\\
Therefore, the appropriate response and explanation for this question is C) Low volatility.

\begin{figure}[ht]
\begin{center}
\includegraphics[width = 1\linewidth]{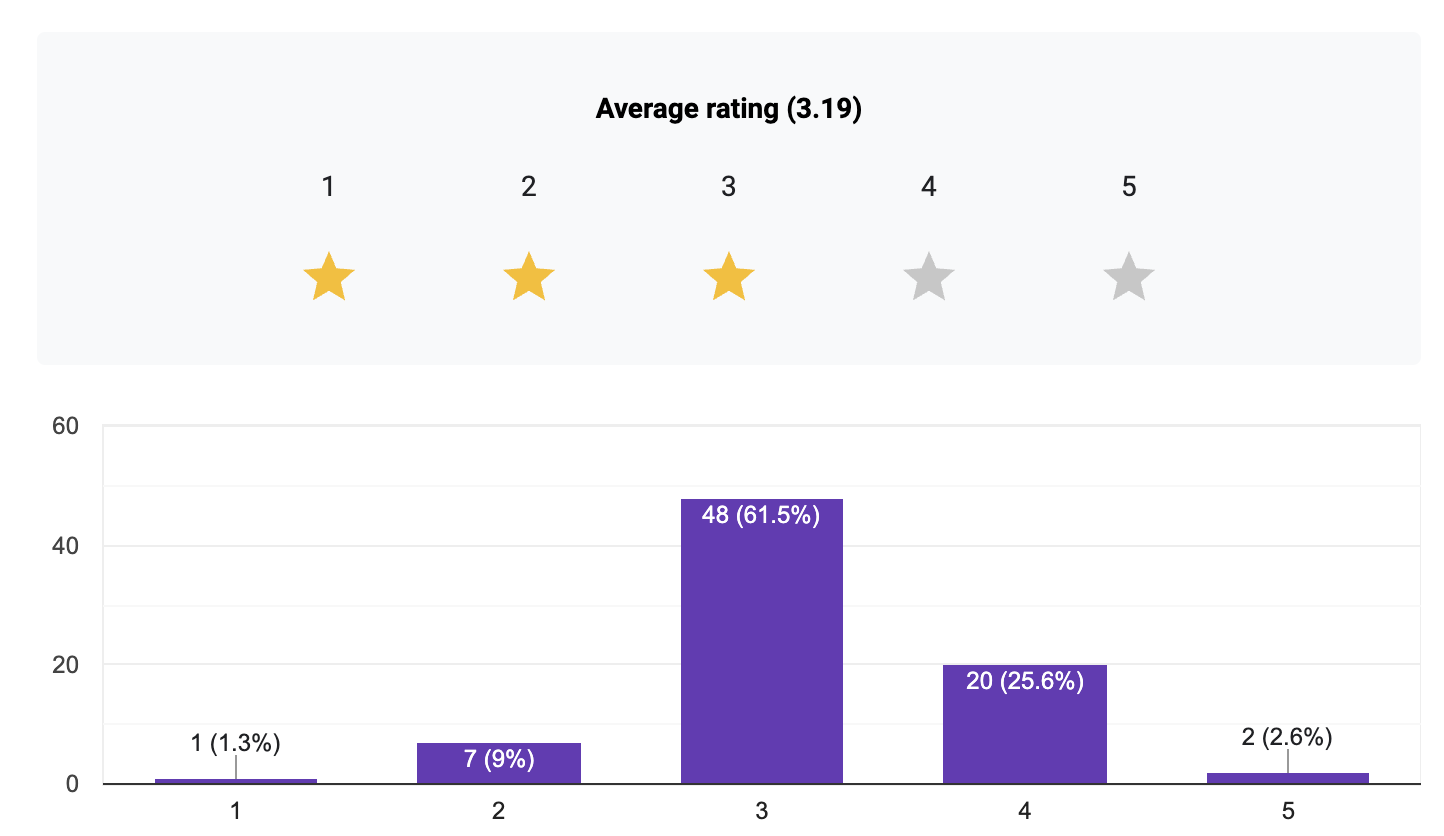}
\end{center}
\end{figure}

\paragraph{Question 9.}
Does the explanation logically follow from the data or the question’s context?
\\~\\
Summarise the overall movement trend in the data points [2.16, 1.86, 1.84, 2.49, 1.98, 1.96, 1.73, 1.79, 2.13, 1.43, 1.49, 1.31, 0.55, -0.16, -0.12, 0.63, 0.66, 0.51, 0.28, -1.14, -2.81, -3.21, -2.26, -1.65].

\begin{figure}[ht]
    \centering
    \begin{subfigure}[b]{\linewidth}
        \includegraphics[width=0.9\linewidth]{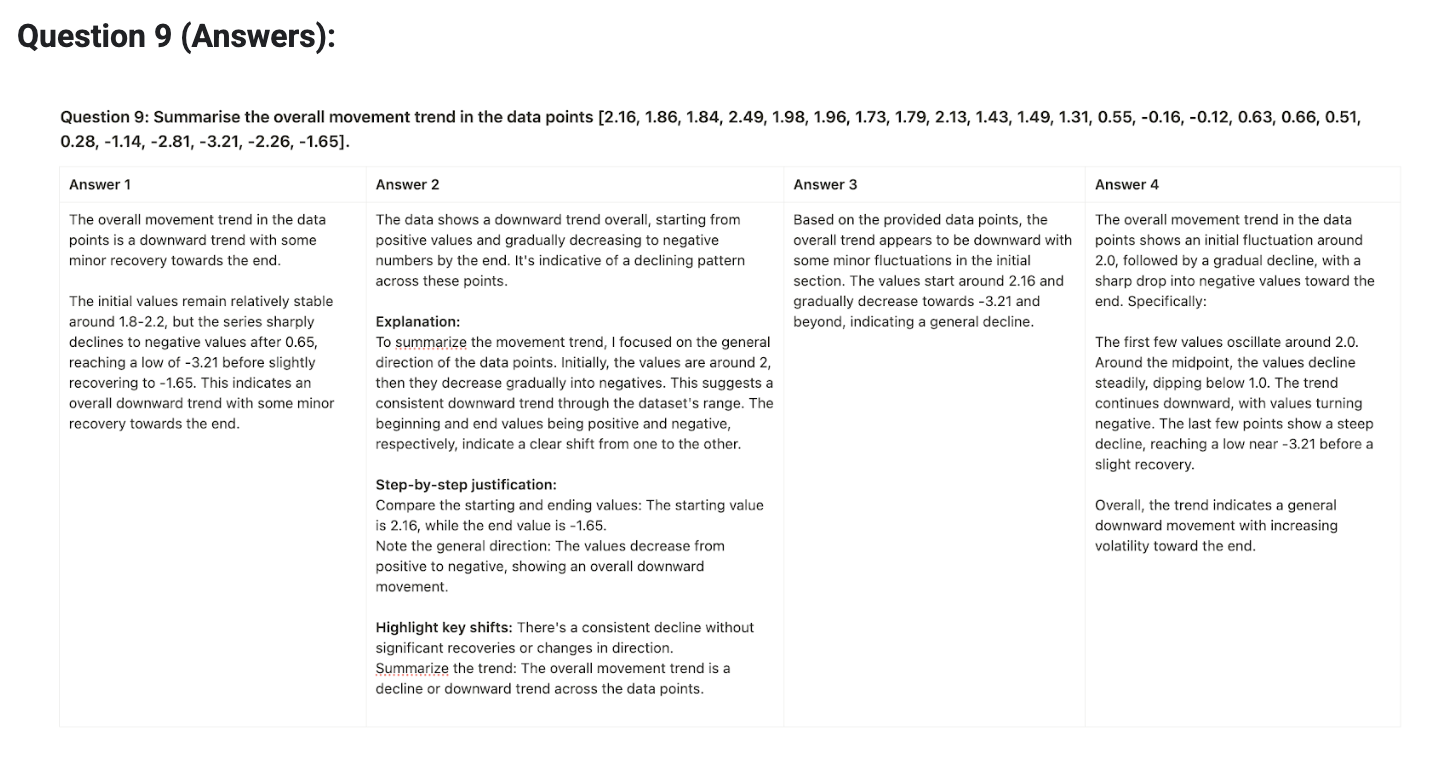}
    \end{subfigure}
    \vspace{0.5cm} 
    \begin{subfigure}[b]{\linewidth}
        \includegraphics[width=0.9\linewidth]{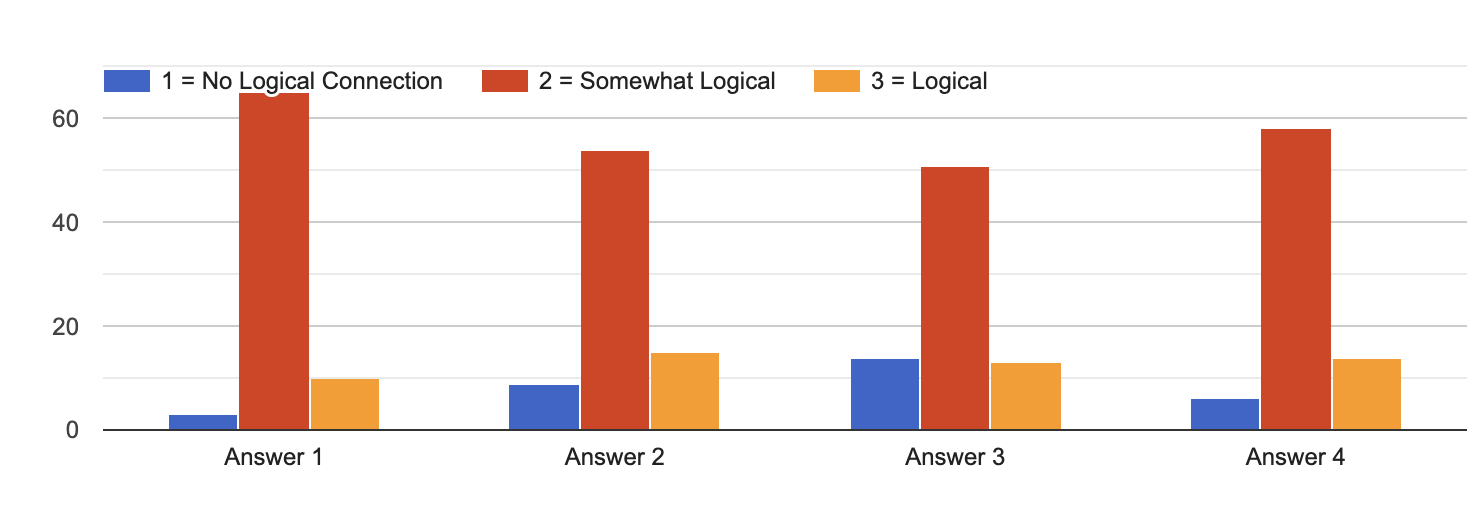}
        \label{fig:q3_2}
    \end{subfigure}
    \vspace{-10mm}
\end{figure}

\paragraph{Question 10.}
How would you rate the quality of the explanation and reasoning with respect to the question? (1 = Really Poor, 5 = Excellent Reasoning)
\\~\\
In the time series data [0.29, 0.29, 0.29, 0.29, 0.29, 0.29, 0.29, 0.29, 0.28, 0.28, 0.28, 0.28, 0.28, 0.28, 0.28, 0.29, 0.3, 0.32, 0.34, 0.37, 0.4, 0.44, 0.47, 0.51], is there evidence of any structural breaks?
\\~\\
Answer: There is a subtle structural break between the stability around 0.28 and the upward shift starting around 0.3.
\\~\\
Explanation: A structural break in time series data is a sudden change in the data pattern or mean level. In the given data, after initially fluctuating around 0.29, there is a persistent drop to 0.28, and then a change occurs towards an upward trend from 0.3 onwards. Although subtle, this transition from 0.28 to 0.3 indicates a structural break where the mean level shifts. It signals the end of one data pattern and the beginning of another distinct phase.
\\~\\
Reasoning: The data stability around 0.29 suggests a constant level, followed by a sudden shift to 0.28, and then an upward trend from 0.3 onwards. This progressive change in behavior indicates there is indeed evidence of a structural break in the sequence.

\begin{figure}[ht]
\begin{center}
\includegraphics[width = 1\linewidth]{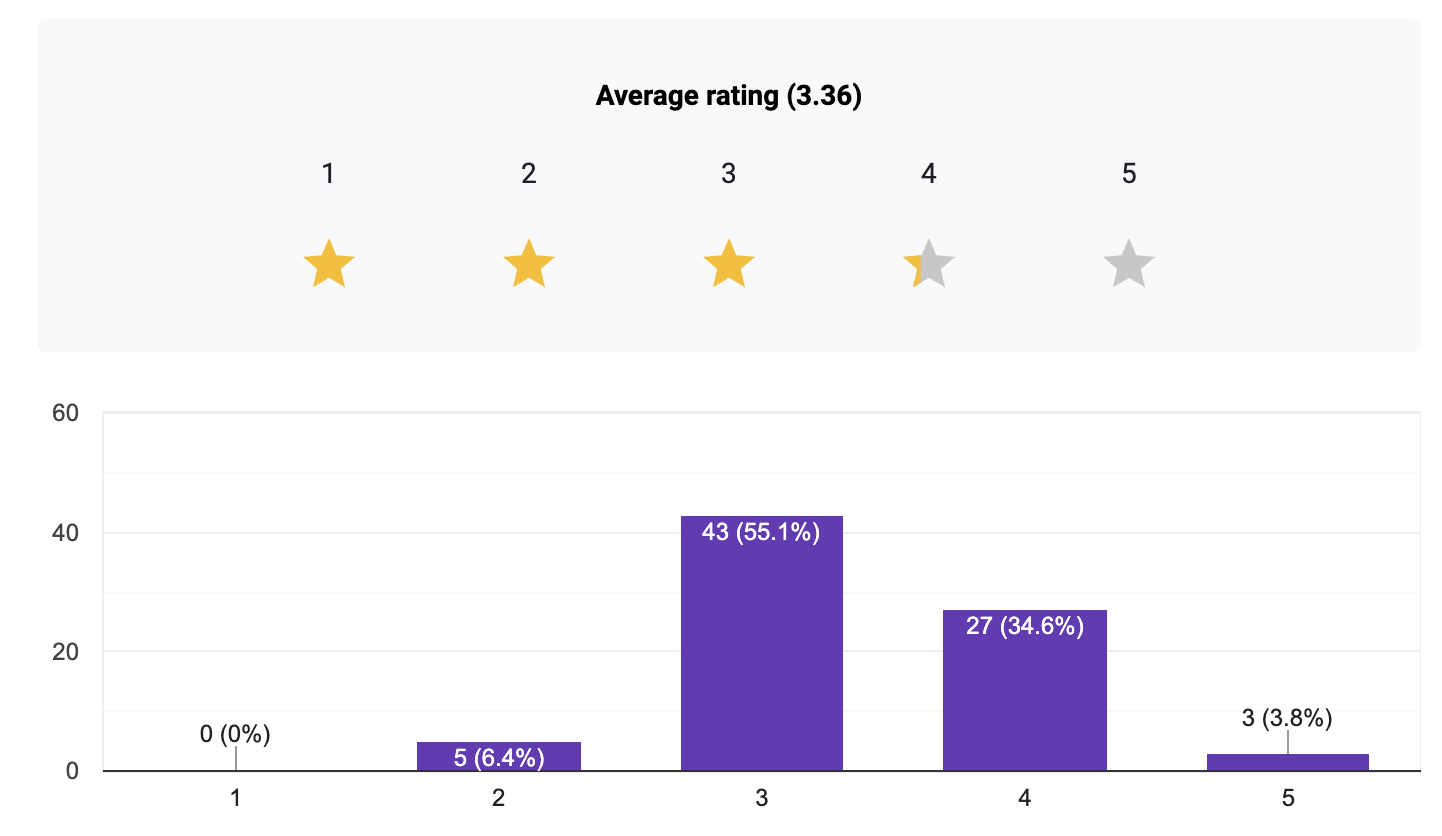}
\end{center}
\end{figure}

\paragraph{Question 11.}
Overall, which answer do you prefer for this question? (You can choose one or more)
\\~\\
Given the data points [6.64, 1.71, 27.49, 1.7, 0.2, 6.05, 33.49, 21.13, 0.0, 0.99, 5.34, 10.74, 2.6, 11.04, 14.36, 9.08, 3.69, 0.26, 8.4, 6.37, 2.35, 6.02, 114.07, 49.47], describe the level of volatility found in this data.

\begin{figure}[ht]
    \centering
    \begin{subfigure}[b]{\linewidth}
        \includegraphics[width=1\linewidth]{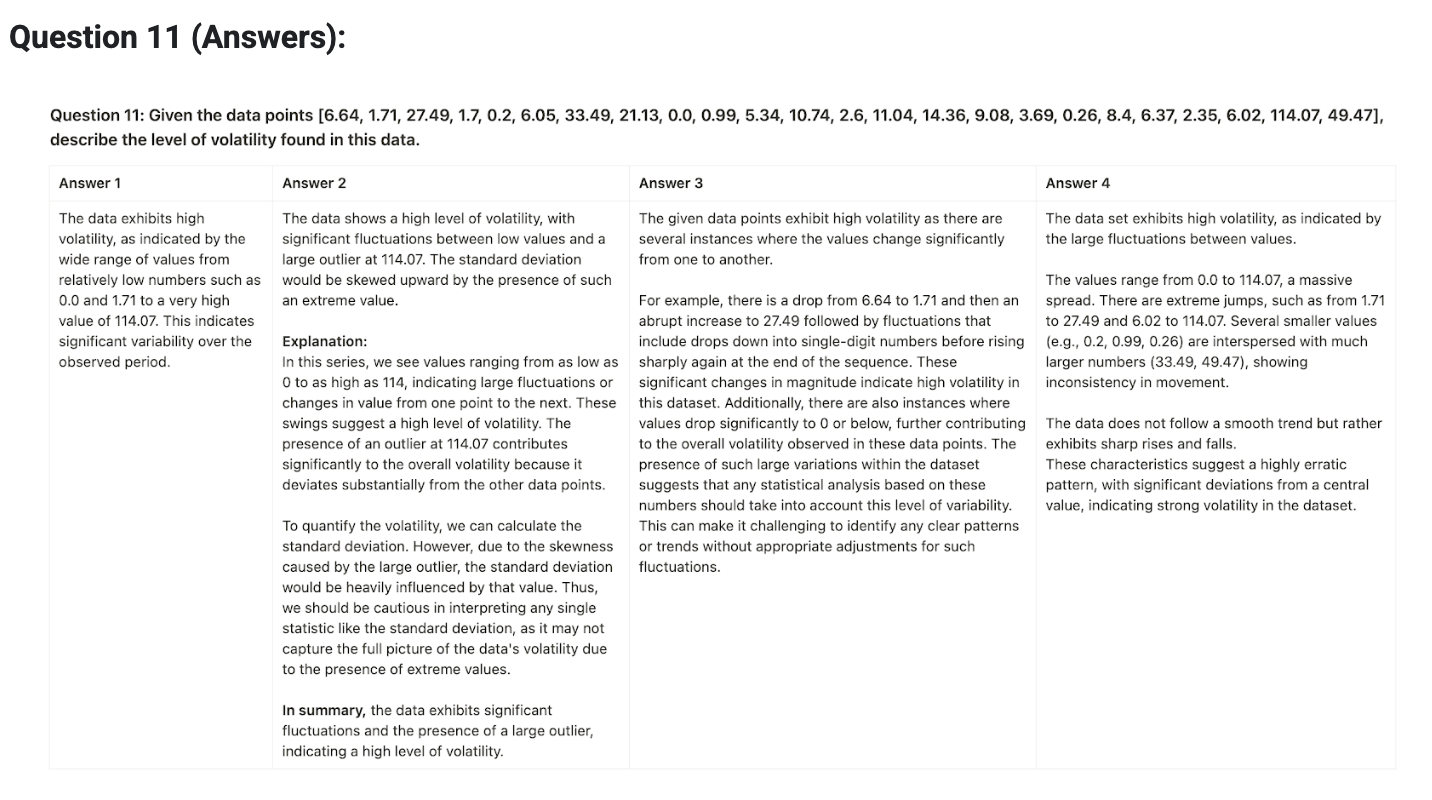}
    \end{subfigure}
    \vspace{0.5cm} 
    \begin{subfigure}[b]{\linewidth}
        \includegraphics[width=1\linewidth]{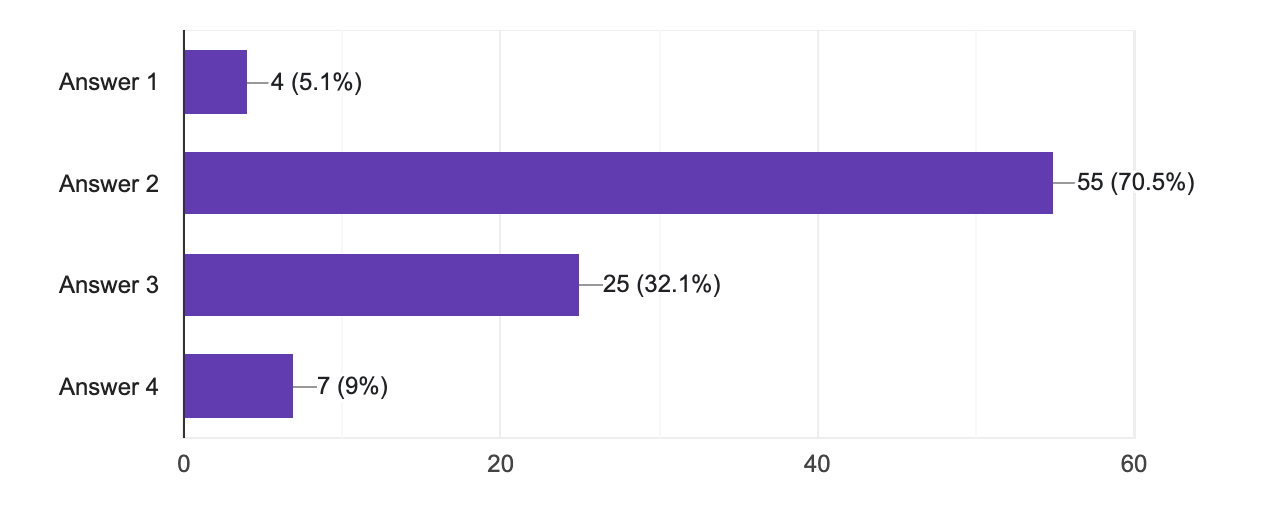}
    \end{subfigure}
    \vspace{-10mm}
\end{figure}

\paragraph{Question 12.}
Which answer do you consider is the best? 
\\~\\
The sequence [0.28, 0.3, 0.34, 0.39, 0.44, 0.49, 0.54, 0.58, 0.62, 0.65, 0.68, 0.7, 0.71, 0.72, 0.73, 0.73, 0.73, 0.73, 0.73, 0.72, 0.71, 0.7, 0.69, 0.68] exhibits an increasing pattern?
\\~\\
Answer 1: The sequence consistently increases until it peaks at 0.73, after which it slightly decreases, but remains relatively high.
\\~\\
Answer 2: The given sequence does not display a clear increasing pattern throughout. While there is an initial increase up to 0.73, followed by a period of stability, the later points decrease consistently from 0.73 to 0.68, breaking the continuity of an upward pattern.
\\~\\
Answer 3: The sequence does show an initial increase from 0.28 to 0.73, but it then stabilizes at 0.73 for several data points before decreasing slightly towards the end (down to 0.68). Therefore, while there is a phase of increase, the overall description suggests both rise and stability/decline, which contrasts with an exclusively increasing pattern. Hence, the correct response is False regarding continuous increase throughout.
\\~\\
Answer 4: The sequence initially increases from 0.28 to 0.73, but after reaching 0.73, it remains constant for a few terms and then starts decreasing from 0.73 to 0.68. Therefore, the sequence does not exhibit a continuously increasing pattern throughout.

\begin{figure}[ht]
\begin{center}
\includegraphics[width = 1\linewidth]{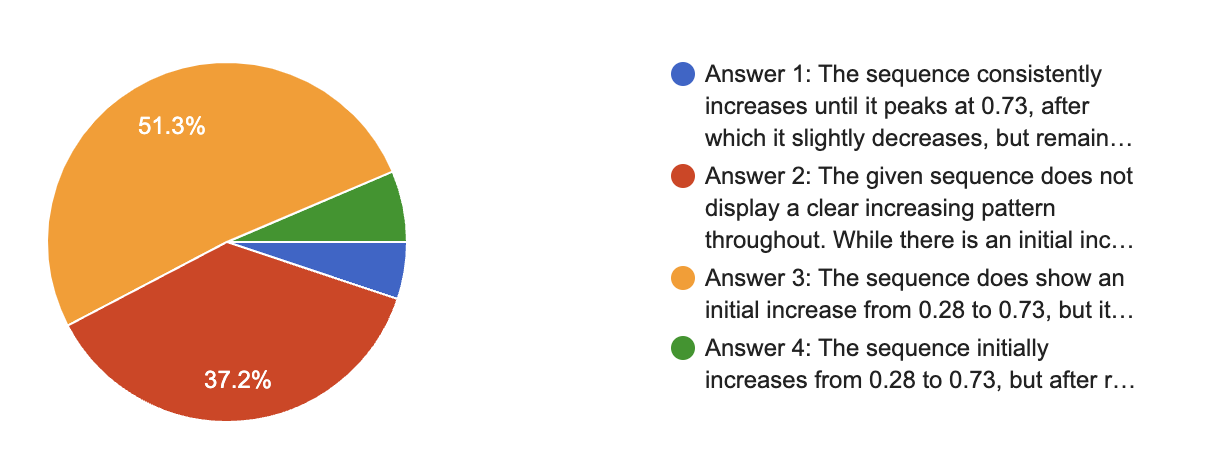}
\end{center}
\end{figure}

\end{document}